\documentclass[letterpaper,journal]{IEEEtran}
\usepackage{amsmath,amsfonts}
\usepackage{algorithmic}
\usepackage{algorithm}
\usepackage{array}
\usepackage[caption=false,font=normalsize,labelfont=sf,textfont=sf]{subfig}
\usepackage{textcomp}
\usepackage{stfloats}
\usepackage{url}
\usepackage{verbatim}
\usepackage{graphicx}
\usepackage{cite}
\usepackage{booktabs}
\usepackage{bm}

\begin{document}

\title{Multimodal Diffusion Transformer with Memory Bank for \\
Scalable Long-Duration Talking Video Generation}

\author{
    Haojie Zhang\textsuperscript{1},
    Zhihao Liang\textsuperscript{1},
    Ruibo Fu\textsuperscript{2,\dag},
    Bingyan Liu\textsuperscript{1},
    Zhengqi Wen\textsuperscript{3},
    Xuefei Liu\textsuperscript{2},\\
    Jianhua Tao\textsuperscript{4},
    Yaling Liang\textsuperscript{1}
    \thanks{\textsuperscript{1}South China University of Technology, Guangzhou, China.}
    \thanks{\textsuperscript{2}Institute of Automation, Chinese Academy of Sciences, Beijing, China.}
    \thanks{\textsuperscript{3}Beijing National Research Center for Information Science and Technology, Tsinghua University, Beijing, China.}
    \thanks{\textsuperscript{4}Department of Automation, BNRist, Tsinghua University, Beijing, China.}
    \thanks{\dag~Corresponding author: ruibofu@126.com}
    \thanks{Code: \protect\url{https://github.com/zhang-haojie/LetsTalk}}
}

\markboth{Zhang \MakeLowercase{\textit{et al.}}: Multimodal Diffusion Transformer with Memory Bank for Scalable Long-Duration Talking Video Generation}%
{Zhang \MakeLowercase{\textit{et al.}}: Multimodal Diffusion Transformer with Memory Bank for Scalable Long-Duration Talking Video Generation}


\maketitle

\begin{abstract}
Long-duration talking video synthesis faces enduring challenges in achieving high video quality, portrait consistency, temporal coherence, and computational efficiency. As video length increases, issues such as visual degradation, portrait drift, temporal artifacts, and error accumulation become increasingly problematic, severely affecting the realism and reliability of the results.
To address these challenges, we present \textbf{LetsTalk}, a diffusion transformer framework equipped with multimodal guidance and a novel memory bank mechanism, explicitly maintaining contextual continuity and enabling robust, high-quality, and efficient generation of long-duration talking videos.
In particular, LetsTalk introduces a noise-regularized memory bank to alleviate error accumulation and sampling artifacts during extended video generation.
To further improve efficiency and spatiotemporal modeling, LetsTalk employs a deep compression autoencoder and a spatiotemporal-aware transformer with linear attention for effective multimodal fusion.
We systematically analyze three fusion schemes and show that combining deep (Symbiotic Fusion) for portrait features and shallow (Direct Fusion) for audio achieves superior visual realism and precise speech-driven motion, while preserving diversity of movements.
Extensive experiments demonstrate that LetsTalk establishes new state-of-the-art in generation quality, producing temporally coherent and realistic talking videos with enhanced diversity and liveliness, and maintains remarkable efficiency with 8$\times$ fewer parameters than previous approaches.
\end{abstract}


\begin{figure*}[!t]
  \centering
  \includegraphics[height=7.0cm]{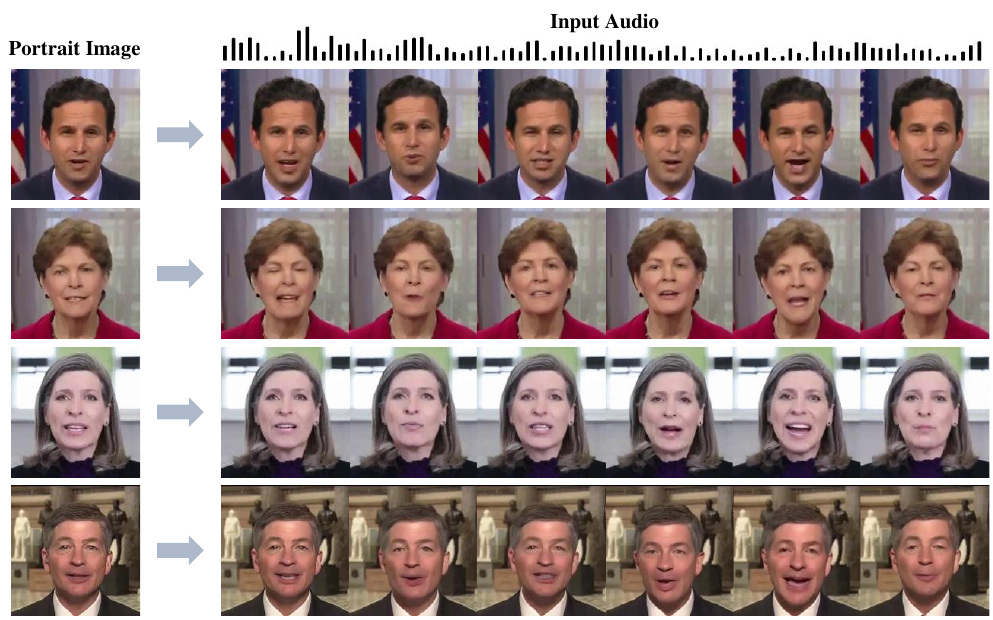}
  \hfil
  \includegraphics[height=7.0cm]{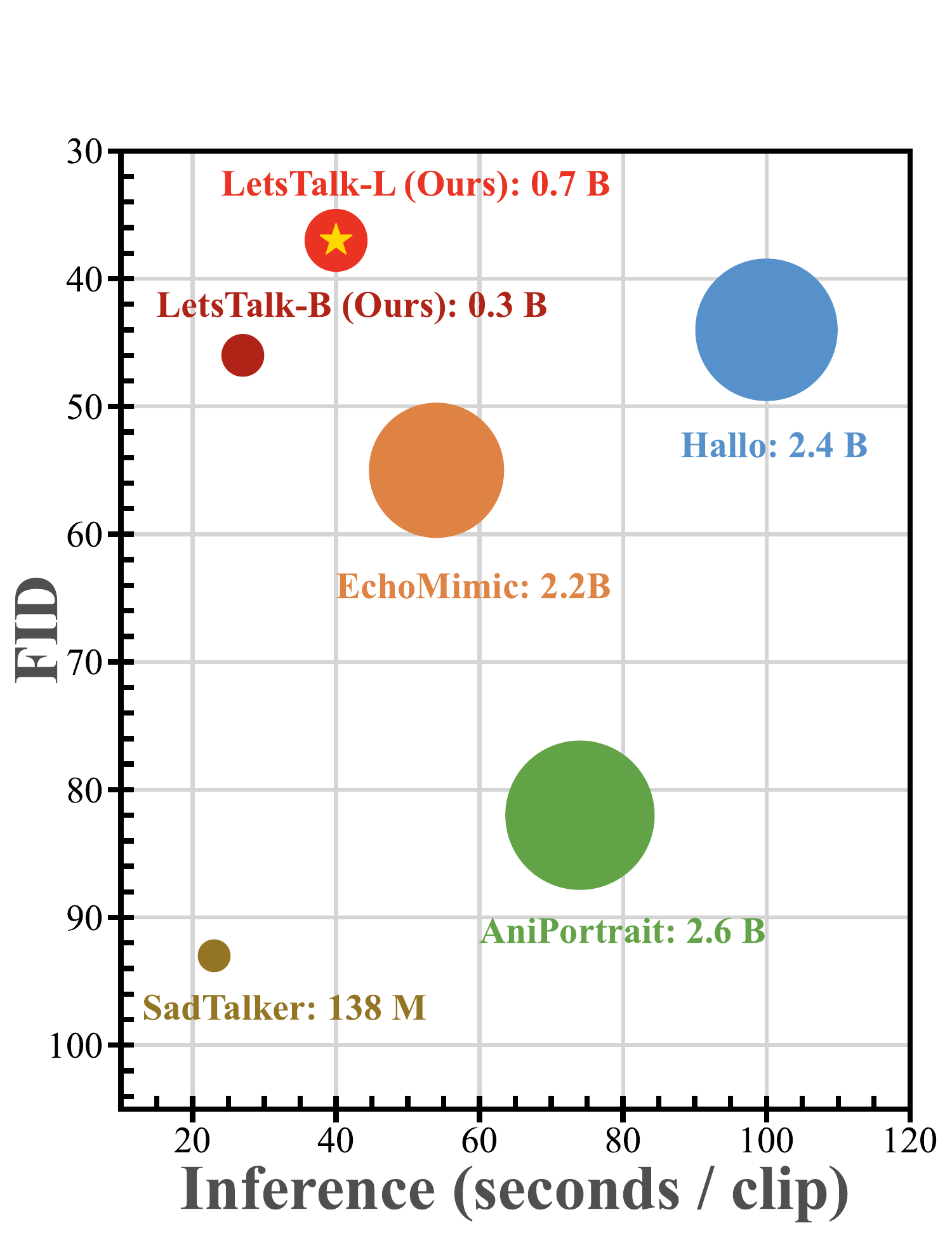}
  \caption{
We introduce \textbf{LetsTalk}, a diffusion-based transformer for audio-driven portrait animation. Given a reference image and audio, LetsTalk generates realistic videos with synchronized mouth motions. As shown in the \textbf{\textit{Left}} figure, each column corresponds to the same audio, demonstrating consistent and accurate lip movements. The \textbf{\textit{Right}} figure compares generation quality and inference time on the HDTF dataset, where circle size represents model parameters. LetsTalk achieves superior quality and efficiency, outperforming methods like Hallo and AniPortrait. Notably, our base version (\emph{LetsTalk-B}) matches Hallo's performance with 8$\times$ fewer parameters.}
  \label{fig:teaser}
\end{figure*}

\section{Introduction}

\IEEEPARstart{A}{nimating} a portrait image with speech, commonly referred to as talking head generation, aims to synthesize a speaking portrait from a single image and accompanying audio. This technology has promising applications in gaming, film production, and digital marketing. Despite recent advancements, long-duration talking video synthesis continues to grapple with maintaining high video quality, portrait consistency, temporal coherence, and computational efficiency~\cite{zhang2023sadtalker, wei2024aniportrait, wang2022anyonenet, ma2025talkclip}. As video duration increases, issues such as visual degradation, portrait drift, and error accumulation become more pronounced, diminishing overall realism and reliability.

Traditional approaches mainly rely on 3D morphable models (3DMM)~\cite{10.1145/311535.311556, li2017learning, bao2021high}, which deliver temporal coherence and stable identity cues but are often limited in their expressiveness~\cite{zhang2023sadtalker, wei2024aniportrait}. Audio-to-2D facial motion techniques~\cite{chen2019hierarchical, prajwal2020lip, zhou2020makelttalk, cheng2022videoretalking, zhou2021pose}, along with multimodal guidance strategies such as Anyonenet~\cite{wang2022anyonenet} and TalkCLIP~\cite{ma2025talkclip}, enrich animation diversity and introduce control via text or voice. Recent progress further enables personalized head movement~\cite{yi2022predicting} and text-audio-guided video generation~\cite{zhao2024ta2v}, but existing methods often struggle to effectively disentangle static and dynamic factors, thereby sometimes compromising portrait consistency or limiting expressiveness across longer time spans.

Diffusion models~\cite{ho2020denoising, dhariwal2021diffusion, rombach2022high} have recently achieved remarkable success in generating realistic 2D images~\cite{rombach2022high, zhang2023adding} and videos~\cite{ma2024latte, lu2023vdt}. AnimateAnyone~\cite{hu2024animate} showcases the strength of motion modules and ReferenceNet for pose-controllable human animation within diffusion frameworks. Building on these advances, recent works~\cite{xu2024hallo, chen2024echomimic, cui2024hallo2} directly employ audio as guidance to synthesize talking head videos, leveraging diffusion models to improve both diversity and expressiveness.

Nevertheless, significant challenges remain for long-duration talking head generation. As video length increases, existing methods frequently struggle to preserve visual quality, portrait consistency, and temporal coherence. Accumulated noise and artifacts, such as distortions and flickering, further undermine realism. Efficiency is often constrained by reliance on auxiliary spatial signals for audio control, unnecessary parameter overhead from fine-tuning generic models for animation tasks, and computationally intensive architectures, including low-compression autoencoders and standard attention schemes. Additionally, effective strategies for fusing multimodal conditions, such as reference portraits and driving audio, have not been sufficiently explored in conditional video generation frameworks.

In this work, we introduce \textit{LetsTalk}, a diffusion transformer framework crafted for efficient, high-quality generation of realistic, long-duration talking videos under multimodal conditions.
A central innovation in LetsTalk for long video generation is the \textit{Noise-Regularized Memory Bank}, which enables each generated clip to access information from prior clips while explicitly simulating inference-time noise accumulation during training. This mechanism effectively preserves contextual continuity and mitigates error accumulation and sampling artifacts throughout extended video sequences, ensuring smooth transitions and strong long-duration consistency.
To further improve efficiency and scalability, LetsTalk incorporates a \textit{Deep Compression Autoencoder} that compresses input frames by 32$\times$ in the latent space, drastically reducing the number of tokens while retaining essential visual features. Transformer blocks are equipped with specialized temporal and spatial modules, utilizing \textit{Linear Attention}~\cite{xie2024sana} for efficient modeling of both intra-frame and inter-frame coherence without excessive computational cost. This design empowers LetsTalk to sustain high generation quality and temporal coherence over long video durations.

For multimodal guidance, defined here as fusing reference portraits and driving audio for video generation, we first categorize three fusion schemes by their integration tightness: 1) \textit{Direct Fusion}, 2) \textit{Siamese Fusion}, and 3) \textit{Symbiotic Fusion}. These schemes are analyzed in terms of integration depth, conditional state, and parameter sharing, providing a systematic summary of their adaptability to diverse modalities. We further examine the underlying factors that influence such adaptability, addressing theoretical gaps in conditional guidance modeling. For effective long-duration video generation, we employ \textit{Symbiotic Fusion} with the reference image to ensure strong portrait consistency across extended sequences due to its visual proximity to the generated video, and use \textit{Direct Fusion} for audio to tightly synchronize lip movement and expressions with speech while fostering diversity in other movements (e.g., blinking, head turns) across longer durations.

We summarize the contributions of this work as follows:
\begin{itemize}
\item We present \textbf{LetsTalk}, a novel and efficient diffusion transformer framework for scalable, long-duration talking video generation, featuring a powerful spatiotemporal modeling architecture that achieves high video quality and computational efficiency with significantly fewer parameters.
\item We propose a noise-regularized memory bank mechanism that preserves contextual continuity and effectively curbs error accumulation, substantially improving long-duration consistency in extended video sequences.
\item We provide a systematic analysis of multimodal fusion schemes and demonstrate that combining deep Symbiotic Fusion for portrait features and Direct Fusion for audio achieves superior visual quality and temporal coherence.
\item Experiments demonstrate that our method achieves realistic, coherent, and diverse long-duration talking videos with optimal quality-efficiency trade-offs.
\end{itemize}

\section{Related Work}

\begin{figure*}[!t]
\centering
\includegraphics[width=0.9\linewidth]{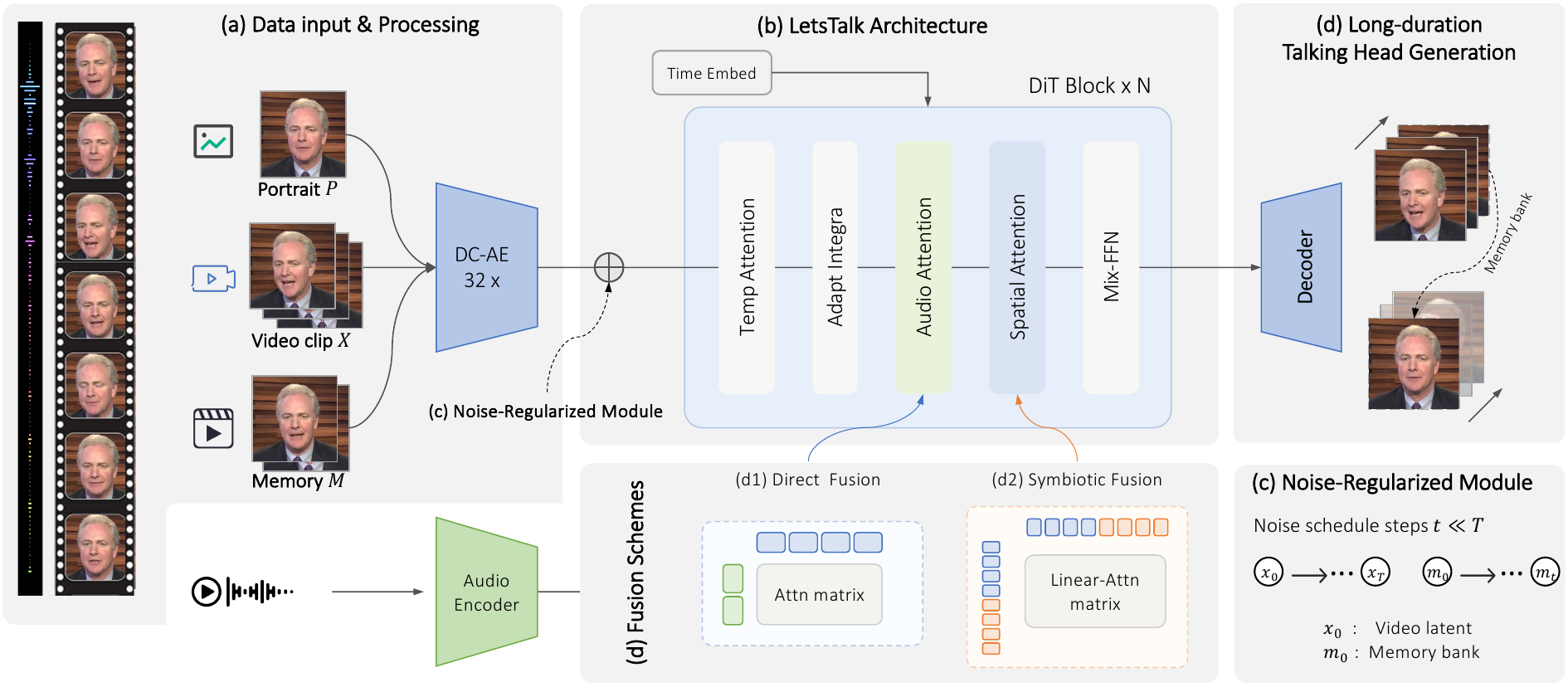}
\caption{
Overview of the proposed \textbf{LetsTalk} framework. (a) During training, the reference portrait $P$, driving video clip $X$, and memory clip $M$ are compressed by a 32$\times$ DC-AE, while the driving audio is encoded by an audio encoder. (b) The denoising backbone is built on stacked DiT blocks with temporal attention, Adapt Integ, audio attention, spatial attention, and Mix-FFN. Portrait features are injected through \textit{Symbiotic Fusion} to preserve portrait appearance, whereas audio features are injected through \textit{Direct Fusion} to provide speech guidance; the bottom insets illustrate the two fusion schemes. (c) To reduce the train-inference gap introduced by memory-based generation, we apply a noise-regularized training strategy that perturbs the memory bank with diffusion noise sampled from early steps ($t \ll T$). (d) During inference, the memory bank stores historical clean features and feeds them back to the decoder, enabling temporally coherent and scalable long-duration talking head generation.
}
\label{fig:overview}
\end{figure*}

\subsection{Video Generation with Diffusion Transformers}
Diffusion models have achieved remarkable success in image synthesis~\cite{ho2020denoising, rombach2022high} and have naturally extended to video generation. Early works like Video Diffusion Models (VDM)~\cite{ho2022video} utilized space-time factorized UNets, while subsequent approaches focused on controllable synthesis by conditioning on text, images, or trajectories~\cite{guo2023animatediff, wang2024videocomposer, chen2023videocrafter1}.
To integrate precise spatial control, methods such as ControlNet~\cite{zhang2023adding} and T2I-Adapter~\cite{mou2024t2i} were introduced, which have been adapted for video to ensure consistency in fine-grained appearance details, exemplified by the ReferenceNet in AnimateAnyone~\cite{hu2024animate}.
A pivotal evolution in this field is the shift from UNet to Diffusion Transformer (DiT) architectures~\cite{peebles2023scalable}, which offer superior scalability. Recent state-of-the-art models, such as Latte~\cite{ma2024latte} and VDT~\cite{lu2023vdt}, adapt DiT for video by incorporating temporal attention mechanisms. Building on this trajectory, our work adopts a DiT-based framework to leverage its efficiency and scalability, while integrating multimodal control (audio and portrait) to achieve high-fidelity generation.

\subsection{Audio-driven Portrait Image Animation}
Portrait image animation has evolved from early methods focusing on basic lip synchronization~\cite{prajwal2020lip, chen2019hierarchical} to incorporating 3D information for more realistic head movements~\cite{zhang2023sadtalker, ye2023geneface}. The field experienced a major breakthrough with the integration of diffusion models, enabling high-fidelity video generation with diverse expressions~\cite{stypulkowski2024diffused, ma2023dreamtalk, sun2023vividtalk}.
Recent innovations have focused on enhancing controllability and expressiveness. Beyond talking-head generation, multimodal generative models such as GALIP~\cite{tao2023galip} and CLIP-GAN~\cite{hou2025clip} demonstrate the effectiveness of CLIP-aligned semantic fusion for controllable generation, highlighting the broader value of multimodal representation alignment. Within portrait animation, approaches like TalkCLIP~\cite{ma2025talkclip} and TA2V~\cite{zhao2024ta2v} introduced natural language guidance for speaking styles. Significant strides in realism were made by works such as EMO~\cite{tian2024emo} and VASA-1~\cite{xu2024vasa}, which focus on highly expressive talking faces. To further refine emotional control, recent work like MoEE~\cite{liu2025moee} introduced a mixture of emotion experts mechanism. The initial Hallo series~\cite{xu2024hallo, cui2024hallo2} improved control over lips and poses using latent diffusion frameworks pioneered in AnimateAnyone~\cite{hu2024animate}. 
Most relevant to our approach is the very recent Hallo3~\cite{cui2025hallo3}, which also shifts from UNet to a Diffusion Transformer architecture, validating the potential of DiT for this task. However, beyond merely adopting a Transformer backbone, our work specifically targets computational efficiency through a linear attention mechanism and deep compression, and introduces a systematic multimodal fusion strategy to optimize the integration of audio and portrait cues for precise animation.

\subsection{Long-Duration Video Generation}
Synthesizing long videos with sustained coherence remains a significant challenge. Traditional 3DMM-based methods~\cite{zhang2023sadtalker} naturally support arbitrary durations but often lack photorealism.
In the diffusion landscape, approaches diverge into tuning-free and training-based strategies. Tuning-free methods, such as ``Global-Local Collaborative Diffusion''~\cite{ma2025tuning} and FreeNoise~\cite{qiu2024freenoise}, employ sliding windows or noise rescheduling to smooth transitions. StreamingT2V~\cite{henschel2025streamingt2v} further enhances this by utilizing short-term memory for autoregressive conditioning. 
Alternatively, training-based solutions like Vlogger~\cite{zhuang2024vlogger} and Hallo2~\cite{cui2024hallo2} optimize specific temporal architectures or augmentation strategies for extended synthesis.
Beyond generation, memory mechanisms in video understanding, specifically the Label Independent Memory~\cite{zhu2022label} for few-shot classification, offer valuable conceptual parallels for retrieving long-term context.
Inspired by these, LetsTalk introduces a trainable Noise-Regularized Memory Bank. Unlike heuristic attention tricks, our approach explicitly simulates and corrects inference-time error accumulation during training, ensuring superior stability for extended sequences.

\section{Methodology}

We propose \textbf{LetsTalk}, a diffusion transformer framework for audio-driven portrait animation, as depicted in Fig.~\ref{fig:overview}. Section~\ref{subsec:backbone} introduces the efficient spatiotemporal transformer backbone. Section~\ref{subsec:long} details our mechanism for long-range consistency, including the memory bank and noise-regularized training. Section~\ref{subsec:fusion} analyzes multimodal fusion schemes, and Section~\ref{subsec:guidance} elaborates our multimodal guidance strategy.

\subsection{Efficient Linear Transformer with Deep Autoencoder}
\label{subsec:backbone}

\noindent\textbf{Deep Compression Autoencoder.} 
We employ a deep compression autoencoder~\cite{xie2024sana} that downsamples images by 32$\times$, achieving reconstruction quality comparable to SDXL while greatly reducing memory and computation costs. Separation of compression and denoising enables preservation of essential temporal cues, such as lip-sync dynamics. Compressed latent features $\mathcal{F} \in \mathbb{R}^{F\times H\times W\times C}$ are partitioned into non-overlapping patches and assigned spatial-temporal positional embeddings~\cite{vaswani2017attention} to enforce structural consistency. This approach facilitates efficient transformer processing and further decreases memory usage by about four times compared to conventional methods.

\noindent\textbf{Interleaved Attention Mechanism.}
Instead of conventional frame-by-frame processing as in DiT~\cite{peebles2023scalable}, we process latent features $\mathbf{Z} \in \mathbb{R}^{F\times P\times d}$ with interleaved spatial and temporal attention, where $F$ denotes the number of frames, $P$ represents the number of spatial patches per frame, and $d$ is the feature dimension.
Full-token attention incurs quadratic complexity $\mathcal{O}(F^2 P^2)$, so we alternate spatial linear attention~\cite{katharopoulos2020transformers} and temporal softmax attention, reducing the total cost to $\mathcal{O}(F^2 + P^2)$. 
Specifically, for spatial modeling, we reshape $\mathbf{Z}$ and apply ReLU-based linear attention across the spatial dimension:
\begin{equation}
    \mathbf{Z}_s = \text{LinearAttn}(\mathbf{Q}_s, \mathbf{K}_s, \mathbf{V}_s) = \frac{\sum_{j=1}^{N} \phi(\mathbf{V}_j) \phi(\mathbf{K}_j)^\top \mathbf{Q}_s}{\sum_{j=1}^{N} \phi(\mathbf{K}_j)^\top \mathbf{Q}_s},
\end{equation}
where $\phi(x) = \text{ReLU}(x)$, and $N=P$ corresponds to the spatial sequence length. 
For temporal modeling, features are transposed to $(P, F, d)$, then subjected to softmax attention across the temporal axis (where the sequence length is $F$). 
This alternating strategy maintains global context and ensures scalability for high-resolution, long-duration inputs.

\noindent\textbf{Mix-FFN Block.}
Although linear attention achieves similar final performance, its convergence may be slower. Following SANA~\cite{xie2024sana}, we replace the standard MLP-FFN with Mix-FFN, which integrates an inverted residual block, a $3\times3$ depth-wise convolution, and a gated linear unit (GLU). This convolution enhances localization ability, compensating for the weaker locality observed with ReLU-based linear attention.

\begin{figure}[t]
\centering
\includegraphics[width=\linewidth]{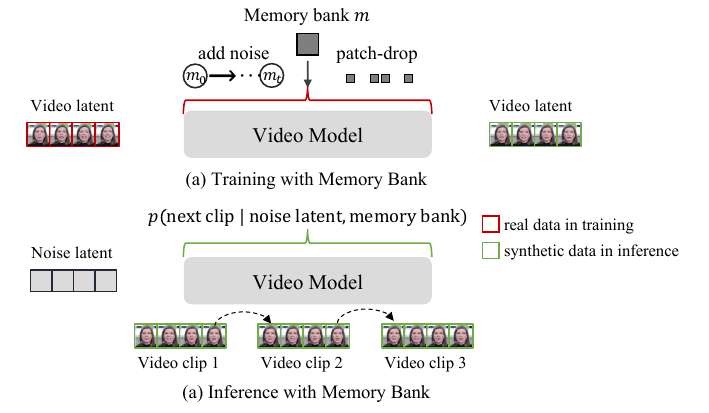}
\caption{The illustration of the noise-regularized memory.}
\label{fig:long-duration}
\end{figure}

\subsection{Long-duration Generation with Noise-Regularized Memory}
\label{subsec:long}
In diffusion-based video generation, maintaining temporal alignment and long-duration consistency across long sequences is essential. However, as the video extends, the accumulation of residual noise and sampling artifacts can threaten temporal coherence and visual quality. This is further exacerbated by the mismatch between training and inference: during training, the memory bank is constructed from ground-truth frames, while during inference, it is updated with generated frames.

\noindent\textbf{Memory Bank:}
To address these challenges, we introduce a \textit{memory bank} that stores information from previously generated clips. Unlike previous frame-by-frame generation approaches~\cite{shen2023difftalk, stypulkowski2024diffused}, our end-to-end diffusion transformer generates an entire video clip in a single sampling process. The memory bank is patchified and temporally transposed, resulting in $\bm{M} \in \mathbb{R}^{P \times f \times d}$, where $f$ denotes the number of frames selected from prior clips. The memory bank provides prior information to the current generation through \textit{Symbiotic Fusion} with $\bm{Z_t}$, leveraging the scalability of the Transformer architecture. This design ensures smooth transitions and long-duration consistency across video segments, and is especially advantageous for long-duration video generation.

\noindent\textbf{Theoretical Motivation:}  
We observed that DDPM-based sampling cannot fully remove noise due to two main factors:
\begin{itemize}
    \item \textit{Model prediction error:} The predicted noise $\epsilon_\theta(x_t, t)$ only approximates the true noise $\epsilon$, leaving some residual noise after denoising.
    \item \textit{Sampling stochasticity:} At each reverse step, new Gaussian noise is added:
    \begin{equation}
        x_{t-1} = \frac{1}{\sqrt{\alpha_t}} \left( x_t - \frac{1-\alpha_t}{\sqrt{1-\bar{\alpha}_t}} \epsilon_\theta(x_t, t) \right) + \sigma_t z \text{.}
    \end{equation}
    The term $\sigma_t z$ introduces inherent randomness.
\end{itemize}
Consequently, even after finite-step denoising, diffusion models retain residual noise, which accumulates as the video sequence extends. This motivates our noise-regularized memory bank, which differs from previous long video generation methods.

\begin{figure}[t]
    \centering
    \includegraphics[width=1.0\linewidth]{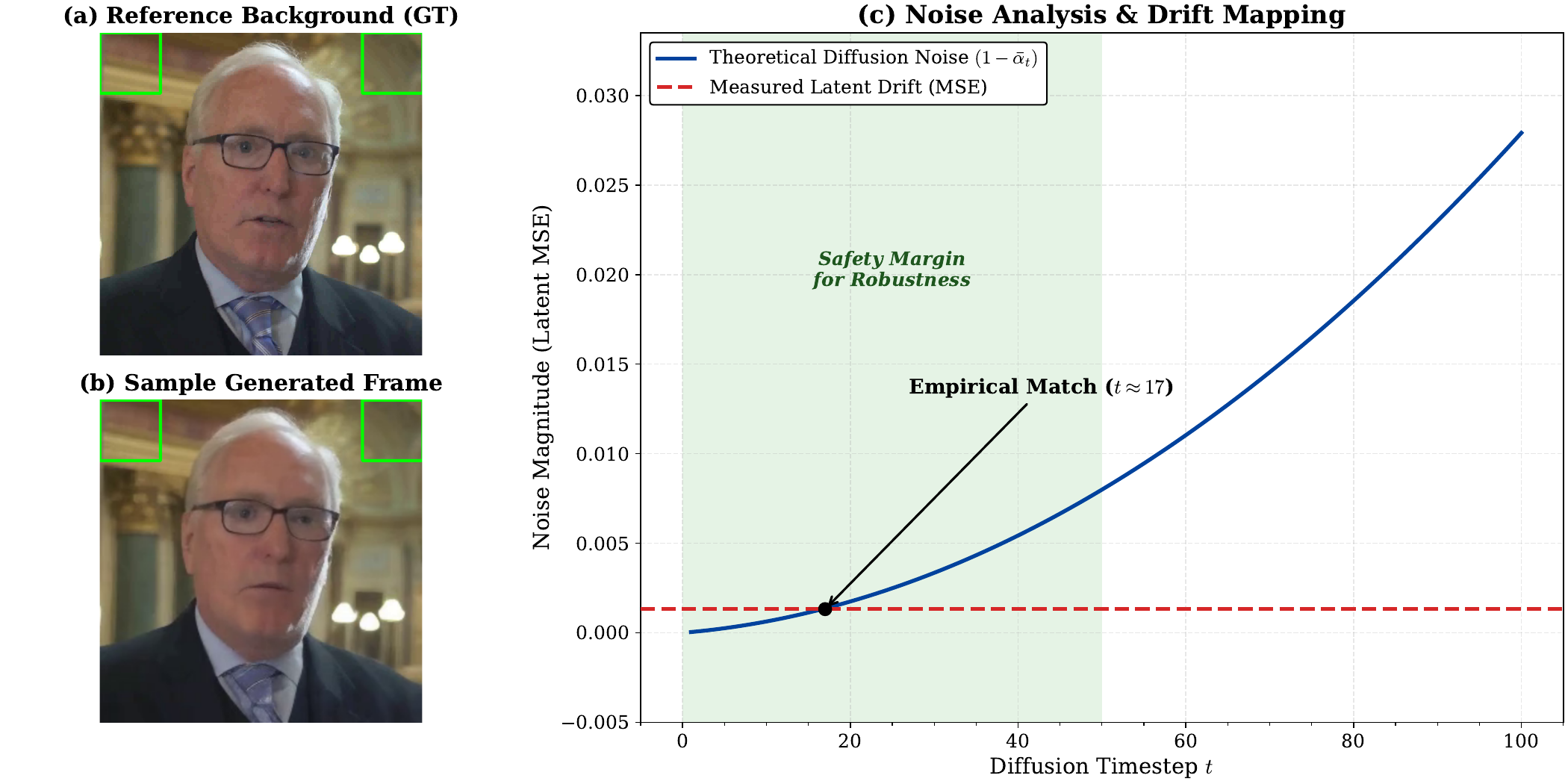} 
    \caption{\textbf{Justification for noise regularization range.} 
    We compare the measured accumulated inference drift (Red Dashed Line) against the theoretical diffusion noise schedule (Blue Solid Curve). 
    The intersection at $t \approx 17$ indicates that our training sampling range $t \in [0, 50]$ (Green Region) sufficiently covers realistic inference errors with a robust safety margin.}
    \label{fig:noise_analysis}
\end{figure}

\noindent\textbf{Noise-Regularized Training Strategy.}
To bridge the gap between training and inference, we propose a noise-regularized training strategy. We perturb the memory bank with controlled Gaussian noise, $\epsilon \sim \mathcal{N}(0, {\sigma_{mini}}^2\mathbf{I})$, simulating inference-time noise accumulation. 

To determine the appropriate noise magnitude, we conducted a quantitative analysis of the residual error distribution during long-duration inference, as shown in Fig.~\ref{fig:noise_analysis}. By measuring the feature drift in static regions, we observed that the actual accumulated noise magnitude aligns with the diffusion noise levels around timestep $t \approx 17$. Based on this observation, we randomly sample a diffusion timestep $t$ from the first 50 steps ($t \in [0, 50]$) and add the corresponding noise to the memory bank during training. This range $[0, 50]$ is chosen to ensure the regularization fully covers the realistic error magnitude while providing a safety margin for potential worst-case deviations (e.g., rapid motion or occlusions), without introducing excessive noise that could degrade feature semantics.

Additionally, inspired by Hallo2~\cite{cui2024hallo2}, we apply patch-drop augmentation by randomly masking 25\% of spatial patches in the memory bank, further enhancing model robustness. This deliberate corruption enables the model to learn to correct for artifacts and maintain temporal dependencies. Our framework thus achieves robust and stable long-duration video synthesis, as demonstrated in Fig.~\ref{fig:long-duration}.

\begin{figure*}[t]
\centering
\includegraphics[height=5cm]{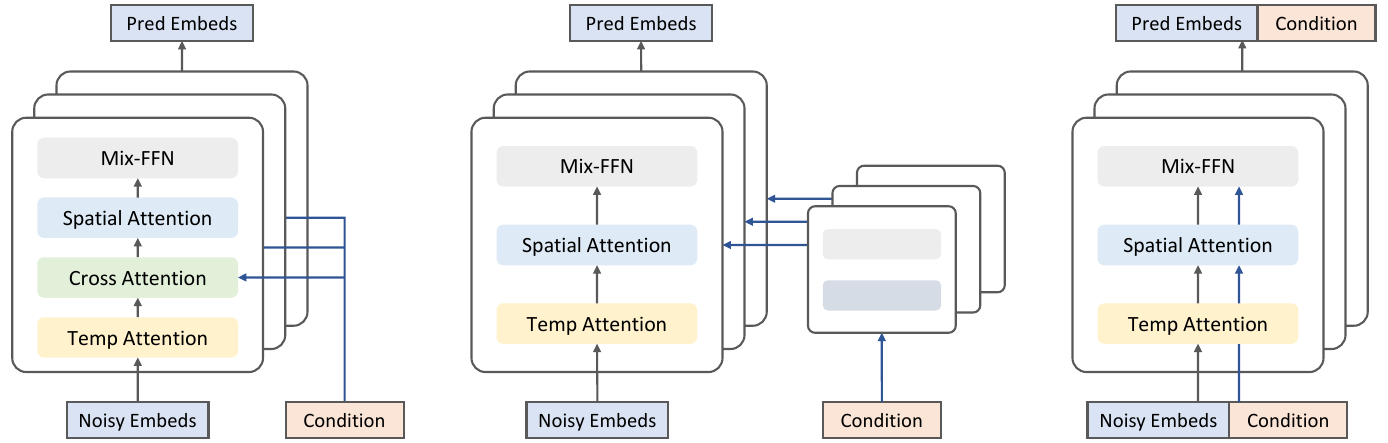}
\caption{Multimodal fusion schemes: (a) \textbf{Direct Fusion} injects conditions via cross-attention modules; 
(b) \textbf{Siamese Fusion} uses parallel transformer for feature guidance; 
(c) \textbf{Symbiotic Fusion} achieves fusion through input concatenation and self-attention. 
The backbone architecture (left-side blocks) remains consistent across all approaches.}
\label{fig:schemes}
\end{figure*}

% \begin{table}[t]
% \centering
% \resizebox{\linewidth}{!}{
% \begin{tabular}{l|ccc|c}
% \toprule
% \textbf{Paradigm} & \textbf{Integration} & \textbf{State} & \textbf{Params} & \textbf{Modality Type} \\ 
% \midrule
% \textbf{Direct}   & Cross-attention    & Static  & Dedicated & Heterogeneous \\ 
% \textbf{Siamese}  & Self-attention/Add & Dynamic & Dedicated & Flexible \\ 
% \textbf{Symbiotic}& Self-attention     & Dynamic & Shared    & Homogeneous \\
% \bottomrule
% \end{tabular}}
% \caption{Systematic categorization of multimodal fusion paradigms. We analyze these schemes to motivate our asymmetric design: using \textbf{Symbiotic} for homogeneous visual references and \textbf{Direct} for heterogeneous audio signals.}
% \label{tab:schemes}
% \end{table}

\begin{table}[t]
\caption{Systematic categorization of multimodal fusion paradigms. We categorize schemes by their architectural inductive biases to motivate our asymmetric design: Symbiotic for portrait consistency and Direct for semantic audio guidance.}
\label{tab:schemes}
\centering
\resizebox{\linewidth}{!}{
\begin{tabular}{l|ccc|c}
\toprule
Paradigm & Integ. & Param. & Align. & Inductive Bias \\
\midrule
Direct    & Cross-attn    & Dedicated & Weak/Static  & Heterogeneous \\ 
Siamese   & Self-attn/Add      & Dedicated & Strong/Spatial & Fine-grained \\ 
Symbiotic & Self-attn     & Shared    & Strong/Pixel & Homogeneous \\
\bottomrule
\end{tabular}}
\end{table}

\subsection{Systematic Analysis and Selection of Fusion Schemes}
\label{subsec:fusion}

To effectively generate portrait videos driven by multimodal inputs such as reference images and audio, selecting fusion mechanisms that align with the specific properties of each modality is essential. Rather than proposing isolated modules, we systematically categorize existing transformer-based fusion paradigms into three primary schemes: Direct, Siamese, and Symbiotic. These are summarized in Table~\ref{tab:schemes}, where we analyze their suitability for our task.
\begin{itemize}
\item \textbf{Direct Fusion}, injects conditions into the backbone via cross-attention. This method serves as the standard conditioning mechanism in text-to-image models like Stable Diffusion~\cite{rombach2022high}. It treats the condition as a static context, keeping the backbone parameters dedicated to the generation target. While computationally efficient, the interaction is limited to the cross-attention bottleneck.
\item \textbf{Siamese Fusion}, utilizes a parallel encoder to extract features from the condition, which are then injected into the backbone via spatial-attention or addition. This architecture is widely adopted by recent methods such as AnimateAnyone~\cite{hu2024animate} (via ReferenceNet), Hallo, and EMO to achieve precise control. However, the use of dedicated parameters for the reference encoder significantly increases model size and training overhead.
\item \textbf{Symbiotic Fusion}, concatenates the condition tokens directly with the latent input, allowing them to interact seamlessly via the backbone's inherent self-attention. Similar strategies are observed in DiffTalk~\cite{shen2023difftalk} and standard DiT architectures. This scheme forces the model to learn a unified representation for both condition and generation, achieving the strongest alignment and parameter efficiency through shared weights.
\end{itemize}

\noindent \textbf{Analysis of Integration and Efficiency.}
The choice among these schemes involves a trade-off between alignment strength and optimization difficulty. Cross-attention (Direct) acts as a semantic bottleneck, making it suitable for bridging heterogeneous modalities with large domain gaps. In contrast, self-attention-based methods (Siamese and Symbiotic) enable pixel-level interaction. Between them, Symbiotic Fusion forces the model to learn a unified representation by sharing weights between the reference and target. This manifold alignment is more parameter-efficient and robust for large-scale training than Siamese Fusion, which requires a decoupled, high-overhead reference encoder (e.g., ReferenceNet) to achieve similar control.

\noindent \textbf{Asymmetric Fusion Design.}
Based on this analysis, we argue that a uniform fusion approach is suboptimal for multimodal inputs with varying properties. We therefore adopt an asymmetric strategy tailored to modality similarity, feature granularity, and optimization stability. For the reference portrait, we use Symbiotic Fusion because the portrait and target frames are both visual tokens with strong spatial correspondence; shared self-attention enables fine-grained portrait consistency without introducing a separate high-overhead branch. For the driving audio, we use Direct Fusion because audio is modality-heterogeneous and mainly provides temporal-semantic guidance rather than pixel-aligned structure. Injecting audio through cross-attention yields a lighter optimization path, preserves the spatial integrity of the visual backbone, and empirically retains more motion diversity than tighter fusion schemes.

\begin{figure*}[!t]
\centering
\includegraphics[width=\linewidth]{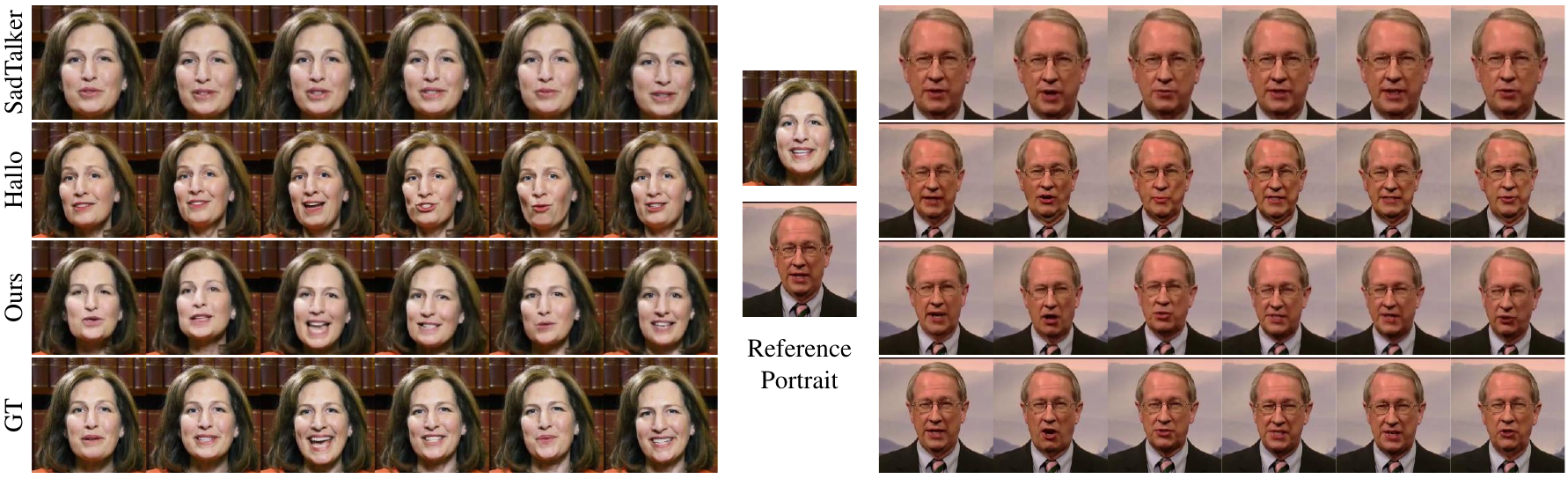}
\caption{
The qualitative comparisons with other cutting-edge methods on the HDTF dataset.
Our method achieves better audio-animation alignment (e.g., lip motions) and produces expressive results.
}
\label{fig:HDTF}
\end{figure*}

\subsection{Guidance on Reference Portrait and Audio}
\label{subsec:guidance}
\noindent\textbf{Symbiotic Fusion for Portrait.}
To enforce strict portrait consistency, we implement Symbiotic Fusion via input concatenation. First, the reference image is encoded by the VAE into a static feature map $\bm{P} \in \mathbb{R}^{h \times w \times c}$. To align with the video generation pipeline, we flatten and temporally replicate $\bm{P}$ across $F$ frames, yielding the sequence $\tilde{\bm{P}} \in \mathbb{R}^{F \times L_p \times d}$.
Crucially, instead of using a separate encoder, we directly concatenate $\tilde{\bm{P}}$ with the noisy video latent features $\bm{Z_s}$ along the spatial dimension. The unified input $\bm{Z} = \text{concat}\left(\bm{Z_s}, \tilde{\bm{P}}\right) \in \mathbb{R}^{F \times (L_s + L_p) \times d}$ is then processed by the transformer backbone. This joint modeling allows the network to attend to reference details via shared self-attention at every layer, ensuring the character's appearance remains robust and consistent throughout the denoising process.

\noindent\textbf{Direct Fusion for Audio.}
For motion guidance, we employ Direct Fusion to handle the heterogeneous audio signals. We extract audio features using a pre-trained wav2vec 2.0~\cite{schneider2019wav2vec} model and project them via a multi-layer perceptron (MLP) into a sequence of tokens $\bm{C} \in \mathbb{R}^{F \times T \times d}$. These tokens are synchronized with the video frames to capture semantic and prosodic cues.
Unlike the portrait features, the audio tokens $\bm{C}$ are injected solely through cross-attention layers inserted after the self-attention blocks. This structural design functions as a conditional bottleneck: it filters the audio information to strictly govern temporal dynamics (e.g., lip-sync and head pose) without interfering with the spatial identity features maintained by the self-attention stream.

\section{Experiment}

\subsection{Experimental Setups}
\noindent\textbf{Datasets.}
We primarily conduct comprehensive experiments on two public datasets: HDTF~\cite{zhang2021flow} and CelebV-HQ~\cite{zhu2022celebv}. These datasets encompass diverse individuals across various ages, ethnicities, and genders, captured in varied environments. To ensure high-quality training data, we follow the preprocessing pipeline in~\cite{xu2024hallo}, excluding videos with scene cuts, extreme camera motion, or side profiles. After refinement, we obtain 356 high-quality clips from HDTF and 24,438 clips from CelebV-HQ, totaling approximately 58.8 hours of footage.

\begin{table}[t]
\caption{Detailed statistics of the dataset splits. We adopt an identity-disjoint strategy, randomly selecting 200 clips for testing to evaluate generalization to unseen identities.}
\label{tab:datasplit}
\centering
\resizebox{0.85\linewidth}{!}{
\begin{tabular}{l|c|cc}
\toprule
{Dataset} & {Total Clips} & {Training Set} & {Test Set} \\ 
\midrule
HDTF~\cite{zhang2021flow} & 356 & 316 & 40 \\ 
CelebV-HQ~\cite{zhu2022celebv} & 24,438 & 24,278 & 160 \\ 
\midrule
{Total} & {24,794} & {24,594} & {200} \\ 
\bottomrule
\end{tabular}
}
\end{table}

\noindent \textbf{Training and Testing Splits.} To rigorously evaluate the model's generalization capabilities to unseen identities, we adopt an identity-disjoint splitting strategy. Specifically, we randomly select 200 distinct video clips (comprising 40 from HDTF and 160 from CelebV-HQ) to form the testing set. The remaining 24,594 clips serve as the training set. This strict separation ensures that the identities in the evaluation phase do not appear during the training process, providing a robust assessment of the model's performance in real-world, open-set scenarios. Detailed statistics of the data distribution are presented in Table~\ref{tab:datasplit}.

\noindent\textbf{Evaluation Metrics.} 
We employ Fréchet Inception Distance (FID) and Fréchet Video Distance (FVD) to assess visual quality and temporal coherence, where lower values indicate higher realism~\cite{karras2019style}. 
To evaluate audio-visual consistency, we utilize the SyncNet scores~\cite{chung2016out}. Sync-C and Sync-D evaluate the lip synchronization of generated videos in terms of content and dynamics, with higher Sync-C and lower Sync-D scores denoting better alignment with the audio.
Additionally, E-FID is used to measure expression fidelity by calculating the feature distance between the extracted expression parameters~\cite{deng2019accurate} of generated and ground truth videos.

\noindent\textbf{Baseline}
In experiments, we compare our method with the publicly available implementations of SadTalker (138M)~\cite{zhang2023sadtalker}, AniPortrait (2.8B)~\cite{wei2024aniportrait}, Hallo (2.4B)~\cite{xu2024hallo}, Hallo 2 (2.4B)~\cite{cui2024hallo2}, Hallo 3 (8B)~\cite{cui2025hallo3} and EchoMimic (2.6B)~\cite{chen2024echomimic}.
The evaluation is performed on the HDTF and CelebV-HQ datasets using the identity-disjoint split described above, with 24,594 training clips and 200 test clips.

\begin{figure*}[t]
\centering
\includegraphics[width=\linewidth]{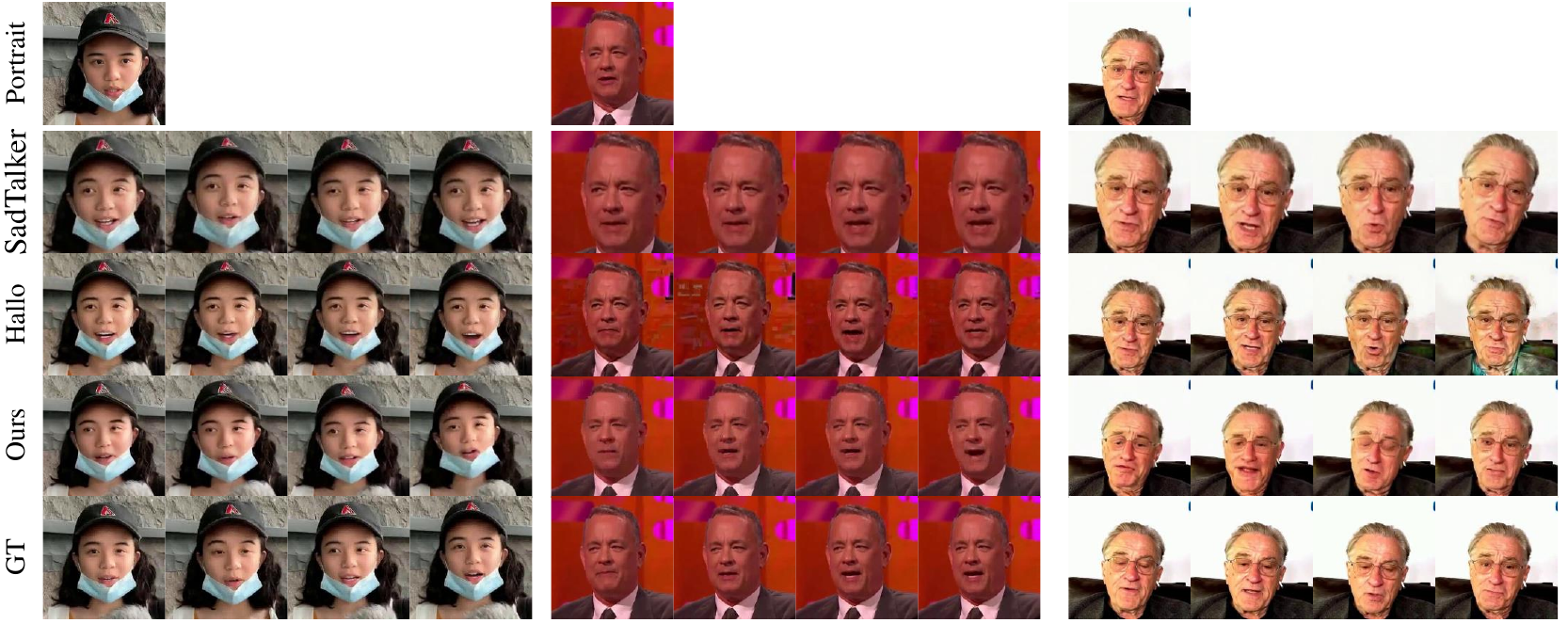}
\caption{The qualitative comparisons with the existing portrait image animation approaches on the CelebV-HQ dataset.
Our method achieves better portrait consistency.
}
\label{fig:CelebV}
\end{figure*}

\begin{table}[t]
\caption{Quantitative comparisons with existing portrait image animation approaches on the HDTF dataset.}
\label{tab:HDTF}
\centering
\resizebox{\linewidth}{!}{
\begin{tabular}{c|cc|cc|c}
\toprule
Method & FID $\downarrow$ & FVD $\downarrow$ & Sync-C $\uparrow$ & Sync-D $\downarrow$ & E-FID $\downarrow$ \\ 
\midrule
SadTalker~\cite{zhang2023sadtalker} & 102.371 & 362.634 & 4.412 & \textbf{7.723} & 2.168 \\ 
AniPortrait~\cite{wei2024aniportrait} & 78.284 & 331.117 & 3.107 & 9.032 & 2.903 \\
EchoMimic~\cite{chen2024echomimic} & 55.791 & 299.343 & 3.398 & 8.987 & 2.493 \\
Hallo~\cite{xu2024hallo} & 45.754 & 285.988 & 4.235 & 8.382 & 2.235 \\
Hallo2~\cite{cui2024hallo2} & 44.343 & 241.628 & 4.482 & 8.121 & 2.017 \\
Hallo3~\cite{cui2025hallo3} & 42.156 & 248.912 & 4.510 & 8.195 & 1.995 \\
\midrule
Ours (Base) & 46.240 & 289.385 & 4.236 & 8.693 & 2.587 \\
Ours (Large) & \textbf{31.761} & \textbf{219.119} & \textbf{4.675} & 8.149 & \textbf{1.962} \\
\bottomrule
\end{tabular}}
\end{table}

\subsection{Additional Implementation Details}
\label{sec:add_implement_details}
Unlike previous methods~\cite{tian2024emo, xu2024hallo, wei2024aniportrait} that rely on multi-stage training, LetsTalk adopts a streamlined end-to-end strategy. We provide two model variants: a Base model for efficient fine-tuning and a Large model for enhanced generalization.

\noindent \textbf{Diffusion \& Training.}
We adopt the hyperparameter setup from ADM~\cite{dhariwal2021diffusion}, utilizing a linear variance schedule with $t_{\text{max}} = 1000$ ranging from $1 \times 10^{-4}$ to $2 \times 10^{-2}$.
The model is initialized from a Latte~\cite{ma2024latte} checkpoint pre-trained on FaceForensics~\cite{rossler2018faceforensics}. We employ the AdamW optimizer with a constant learning rate of $1 \times 10^{-4}$ and no weight decay. Training runs for 300k to 500k iterations on NVIDIA RTX 3090 and A800 GPUs using automatic mixed precision. We apply gradient clipping and horizontal flip augmentation for stability, and maintain an exponential moving average (EMA) of weights (decay rate $0.9999$) for final evaluation.

\noindent\textbf{Inference.}
At inference, the model generates a video sequence from a single reference portrait and driving audio. Consistent with training, we generate 16-frame clips at $512\times 512$ resolution. To ensure high generation quality, we utilize the DDPM sampling strategy with $250$ steps.

\subsection{Quantitative Results}
We report comprehensive quantitative comparisons on the HDTF and CelebV-HQ datasets in Tables~\ref{tab:HDTF} and~\ref{tab:CelebV}, respectively. As shown in the tables, our method achieves the best performance on most metrics. Specifically, on the HDTF dataset, LetsTalk-L achieves the lowest FID (31.761) and FVD (219.119), outperforming the previous best method Hallo2 by 28.3\% and 9.3\%, respectively. On CelebV-HQ, our method also achieves the lowest FID (7.610) and FVD (33.175). These results demonstrate that our approach generates more realistic and temporally consistent talking-head videos. All methods generate the first video clip per validation sample and are evaluated at the same resolution.

\begin{table}[t]
\caption{The quantitative comparisons with the existing portrait image animation approaches on the CelebV-HQ dataset.}
\label{tab:CelebV}
\centering
\resizebox{\linewidth}{!}{
\begin{tabular}{c|cc|cc|c}
\toprule
Method & FID $\downarrow$ & FVD $\downarrow$ & Sync-C $\uparrow$ & Sync-D $\downarrow$ & E-FID $\downarrow$ \\ 
\midrule
SadTalker~\cite{zhang2023sadtalker} & 42.017 & 77.802 & 4.150 & 8.423 & 3.372 \\ 
AniPortrait~\cite{wei2024aniportrait} & 33.408 & 107.236 & 2.955 & 9.680 & 3.722 \\
EchoMimic~\cite{chen2024echomimic} & 12.932 & 59.944 & 3.120 & 9.350 & 3.671 \\
Hallo~\cite{xu2024hallo} & 8.504 & 46.431 & 3.850 & 8.840 & 3.280 \\
Hallo2~\cite{cui2024hallo2} & 8.734 & 41.213 & 4.050 & 8.710 & 3.128 \\
Hallo3~\cite{cui2025hallo3} & 8.321 & 41.550 & 4.120 & 8.650 & \textbf{3.012} \\
\midrule
Ours (Base) &  9.585 & 42.213  & 3.920 & 8.950 & 3.349 \\
Ours (Large) & \textbf{7.610} & \textbf{33.175} & \textbf{4.385} & \textbf{8.350} & 3.038 \\
\bottomrule
\end{tabular}}
\end{table}

\subsection{Qualitative Results}
We have visualized the results of different methods on the HDTF dataset (Fig.~\ref{fig:HDTF}) and the CelebV-HQ dataset (Fig.~\ref{fig:CelebV}).
From the visualization results, it is evident that our method performs exceptionally well on both datasets.
In comparison to the visual cues most strongly associated with audio (such as facial muscle and lip motions), our results closely align with the standard Ground Truth.
Benefitting from our \textit{Symbiotic Fusion}, our generated videos maintain strict portrait consistency.
Furthermore,
by employing \textit{Direct Fusion} of audio, the characters exhibit strong expressiveness and diversity in our generated videos while achieving audio-animation alignment.

\begin{table}[t]
\caption{Comparison of model size, inference time (ms per frame), and FID on the HDTF dataset.}
\label{tab:efficiency}
\centering
\resizebox{0.9\linewidth}{!}{
\begin{tabular}{l|c|c|c}
\toprule
Method & \# Params & Latency (ms) & FID $\downarrow$ \\
\midrule
SadTalker~\cite{zhang2023sadtalker} & 138M & 23 & 102.371 \\
EchoMimic~\cite{chen2024echomimic} & 2.2B & 54 & 55.791 \\
AniPortrait~\cite{wei2024aniportrait} & 2.6B & 74 & 78.284 \\
Hallo~\cite{xu2024hallo} & 2.4B & 115 & 44.343 \\
\midrule
LetsTalk-B (Ours) & 0.3B & \underline{27} & 46.240 \\
LetsTalk-L (Ours) & 0.7B & 40 & \textbf{31.761} \\
\bottomrule
\end{tabular}
}
\end{table}

\begin{figure*}[t]
  \centering
  \includegraphics[width=\linewidth]{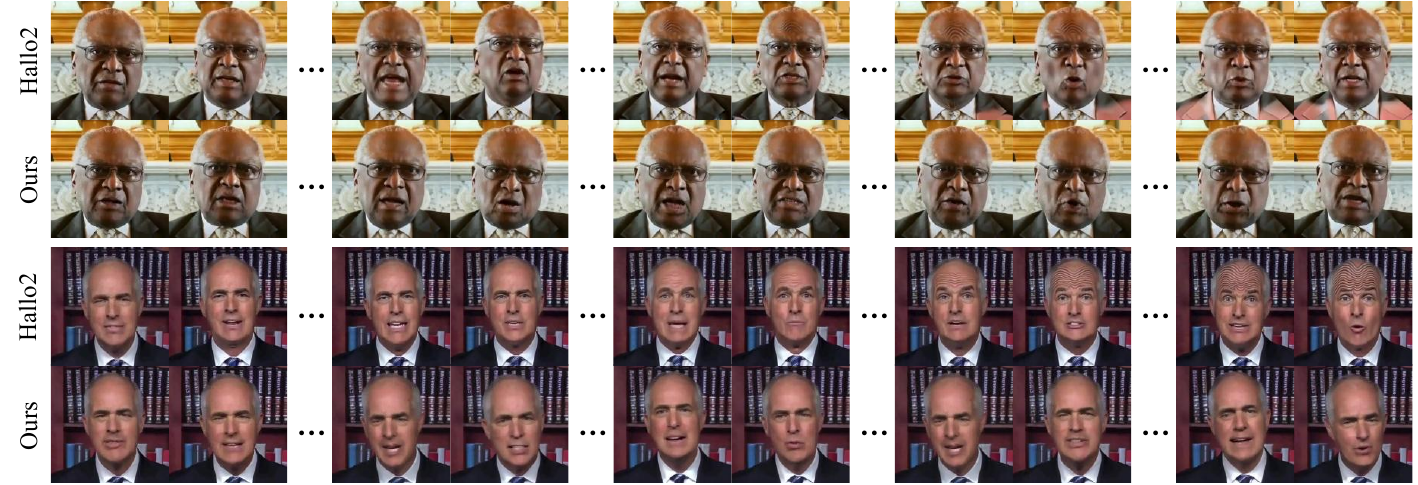}
  \caption{Long-duration qualitative comparison with Hallo2. We visualized frames sampled from a continuous 4-minute generation (0 to approx. 6,800 frames). While Hallo2 exhibits obvious error accumulation artifacts (e.g., forehead textures) in later stages, our method maintains consistent visual quality and identity stability.}
  \label{fig:long_duration_comparison}
\end{figure*}

\subsection{Long-Duration Generation Assessment}

To validate the stability of our model in long-duration scenarios and the effectiveness of the proposed Memory Bank in mitigating error accumulation, we conducted both quantitative ablation studies and qualitative comparisons with state-of-the-art methods.

% \noindent \textbf{Ablation on Components.} We first evaluated our approach on 100 consecutive clips quantitatively. As shown in \cref{tab:long_video_results}, Noise-Regularized Training plays a crucial role in visual quality, achieving the best FID (32.746). However, for long-duration consistency, temporal coherence is paramount. Although adding Patch Drop yields a slightly higher FID (34.821), it achieves the lowest FVD (251.684). This indicates that Patch Drop introduces minor spatial noise to enforce stronger reliance on temporal context, thereby smoothing transitions. Consequently, our final model employs both strategies to balance fidelity and consistency.

\noindent \textbf{Ablation on Components.} We first evaluated our approach on 100 consecutive clips quantitatively. Since the Memory Bank is the key mechanism for cross-clip context propagation in our long-duration setting, we keep it enabled in all variants and focus this ablation on the effects of Noise-Regularized Training and Patch Drop. As shown in Table~\ref{tab:long_video_results}, Noise-Regularized Training plays a crucial role in visual quality, achieving the best FID (32.746). However, for long-duration consistency, temporal coherence is paramount. Although adding Patch Drop yields a slightly higher FID (34.821), it achieves the lowest FVD (251.684). This indicates that Patch Drop introduces minor spatial noise to enforce stronger reliance on temporal context, thereby smoothing transitions. Consequently, our final model employs both strategies to balance fidelity and consistency.

\noindent \textbf{Comparison with State-of-the-Art.} To further verify the robustness against error accumulation, we conducted a rigorous qualitative comparison with Hallo2, a leading method designed for long-duration generation. We generated video sequences extending beyond 4 minutes (approx. 6,800 frames at 25 fps). 
As illustrated in Fig.~\ref{fig:long_duration_comparison}, while Hallo2 maintains quality in the initial stages, it suffers from severe degradation as generation proceeds. Noticeable artifacts, such as unnatural texture interference on the forehead and identity distortion, appear in later frames (e.g., towards frame 6800), indicating significant error accumulation. 
In contrast, thanks to the explicit context retrieval provided by the Memory Bank, our method maintains consistent visual fidelity and portrait consistency throughout the entire sequence, exhibiting no signs of temporal drift or artifact accumulation.

\begin{figure}[t]
\centering
\includegraphics[width=\linewidth]{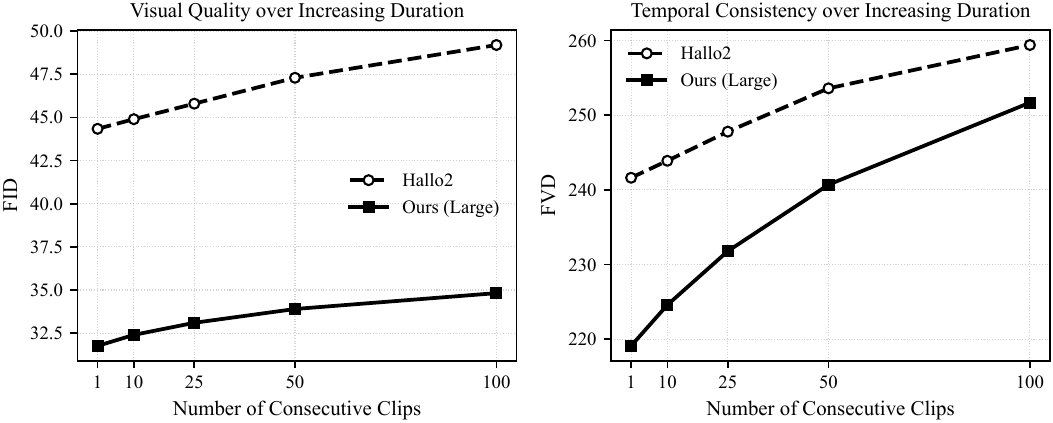}
\caption{Long-duration consistency over increasing generation horizons. We report FID and FVD as the number of consecutively generated clips increases. Although both methods degrade gradually over longer horizons, LetsTalk-L exhibits a consistently slower performance drop than Hallo2, demonstrating stronger robustness against long-duration error accumulation.}
\label{fig:temporal_consistency_curves}
\end{figure}

\noindent \textbf{Long-Duration Consistency over Increasing Duration.} To provide a more direct quantitative view of long-duration stability, we further evaluate FID and FVD as the number of consecutively generated clips increases. As shown in Fig.~\ref{fig:temporal_consistency_curves}, both methods exhibit gradual degradation as the generation horizon becomes longer. However, the increase in FID and FVD is consistently slower for LetsTalk-L than for Hallo2, indicating that our method better suppresses long-duration error accumulation while preserving visual fidelity. This result is consistent with the qualitative comparison in Fig.~\ref{fig:long_duration_comparison} and further supports the effectiveness of the proposed Memory Bank for maintaining long-duration consistency over extended generation.

\begin{table}[t]
\caption{Quantitative ablation results on long video generation (100 clips). \checkmark\ indicates the component is enabled.}
\label{tab:long_video_results}
\centering
\resizebox{\linewidth}{!}{%
\begin{tabular}{ccc|cc}
\toprule
Memory Bank & Patch Drop & Regularized Training & FID$\downarrow$ & FVD$\downarrow$ \\
\midrule
\checkmark &             &                       & 52.243 & 276.16 \\
\checkmark & \checkmark &                       & 43.728 & 263.173 \\
\checkmark &             & \checkmark            & \textbf{32.746} & 264.709 \\
\checkmark & \checkmark & \checkmark            & 34.821 & \textbf{251.684} \\
\bottomrule
\end{tabular}%
}
\end{table}

\subsection{Efficiency Comparison}
To further demonstrate the practical value of our method, we compare model size, inference speed, and FID score with previous state-of-the-art methods. As shown in Table~\ref{tab:efficiency}, LetsTalk-B achieves comparable or better visual quality with significantly fewer parameters and faster inference than recent diffusion-based approaches. The larger LetsTalk-L further improves FID while maintaining reasonable efficiency. 
These results highlight the strong trade-off between quality and efficiency achieved by LetsTalk, making it well-suited for real-world applications.

\subsection{Ablation Study}
\label{sec:ablation}
In this section, we analyze the impact of different fusion schemes as illustrated in Fig.~\ref{fig:schemes} on guidance using reference portrait and audio.

\begin{table}[t]
\caption{Experimental results of three multimodal fusion schemes, guided by reference portraits, on the HDTF validation dataset, employing the LetsTalk-B model.}
\label{tab:abla_portrait}
\centering
\resizebox{\linewidth}{!}
{
\begin{tabular}{l|ccccc}
\toprule
Reference Portrait & FID $\downarrow$ & FVD $\downarrow$ & Sync-C $\uparrow$ & Sync-D $\downarrow$ & E-FID $\downarrow$ \\ 
\midrule
Direct Fusion & 52.999 & 389.01 & 3.652 & 9.421 & 3.149 \\ 
Siamese Fusion & 47.779 & \textbf{286.692} & 4.015 & 8.925 & 2.597 \\ 
Symbiotic Fusion & \textbf{46.240} & 289.385 & \textbf{4.236} & \textbf{8.693} & \textbf{2.587}  \\
\bottomrule
\end{tabular}
}
\end{table}

\begin{table}[t]
\caption{Experimental results of three multimodal fusion schemes, driven by audio inputs, on the HDTF validation dataset, employing the LetsTalk-B model.}
\label{tab:abla_audio}
\centering
\resizebox{\linewidth}{!}
{
\begin{tabular}{l|ccccc} 
\toprule
Driven Audio & FID $\downarrow$ & FVD $\downarrow$ & Sync-C $\uparrow$ & Sync-D $\downarrow$ & E-FID $\downarrow$ \\ 
\midrule
Direct Fusion & \textbf{46.240} & \textbf{289.385} & \textbf{4.236} & \textbf{8.693} & \textbf{2.587} \\ 
Siamese Fusion & 48.041 & 318.637 & 3.845 & 9.102 & 2.706 \\ 
Symbiotic Fusion & 51.880 & 403.628 & 3.421 & 9.854 & 3.398 \\
\bottomrule
\end{tabular}
}
\end{table}

\noindent\textbf{Analysis on the Reference Portrait Fusion}
Portrait consistency is crucial to the portrait image animation task and highly relevant to the portrait fusion scheme used.
We evaluate the synthesis quality of portrait fusion using three schemes, and the quantitative results are shown in Table~\ref{tab:abla_portrait}.
\textit{Symbiotic Fusion} almost surpasses the other two schemes, realizing better portrait consistency and video quality.
Besides, while \textit{Siamese Fusion} produces comparable results, it introduces additional parameters (\emph{i.e.} siamese transformer) and complicates the learning task.

\noindent\textbf{Analysis on the Driven Audio Fusion}
The way we fuse audio affects both audio-animation alignment and diversity.
Audio-animation alignment refers to specific visual cues~\cite{zhang2023sadtalker} (e.g., facial muscle and lip motions), whereas the diversity is related to the speech-agnostic motions (e.g., blinking and neck moving) that often appear as unconscious actions.
We observe that diversity mainly comes from the transformer backbone during denoising.
Thus, we aim to fuse the audio to achieve alignment while preserving the backbone's inherent diversity.
Quantitative results in Table~\ref{tab:abla_audio} indicate that \textit{Direct Fusion} outperforms the other schemes in establishing the correspondence between audio and animation.
Taken together with the results in Table~\ref{tab:abla_portrait}, these observations further support our asymmetric design choice: \textit{Symbiotic Fusion} is more suitable for the reference portrait, where tighter visual interaction benefits portrait consistency, whereas \textit{Direct Fusion} is more effective for the driving audio, where explicit conditioning better preserves audio-animation alignment and motion diversity.

\begin{figure}[!t]
\centering
\includegraphics[width=\linewidth]{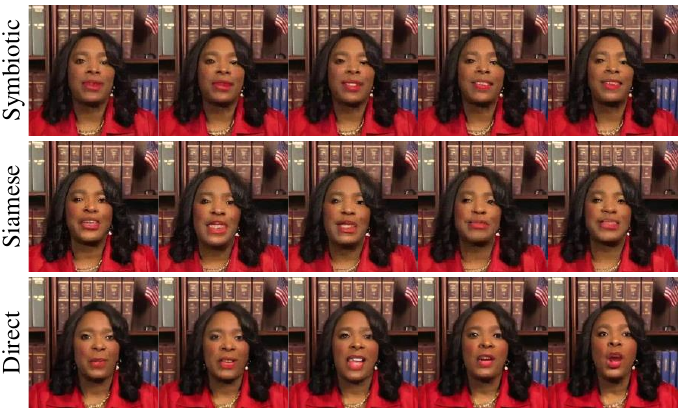}
\caption{Portrait animation results without audio guidance for three fusion schemes.
Siamese and Symbiotic fusion schemes with high fusion compactness limit the diversity of video generation, while Direct fusion preserves this diversity.
}
\label{fig:Audio}
\end{figure}

\noindent\textbf{Impact of Audio-guidance on Diversity}
To validate the impact of different fusion schemes on diversity, we serve LetsTalk as an unconditional video generator\footnote{Filling audio embeddings with $0$ for fusion.} and compare the diversity of character actions in the generated videos.
The results in Fig.~\ref{fig:Audio} demonstrate that the deeper the fusion compactness, the more static the generated video.
The tightest \textit{Symbiotic Fusion} generates a portrait with almost no motions, \textit{Siamese Fusion} produces slight motions, while the least tight \textit{Direct Fusion} we adopt results in natural motions.

\begin{figure}[t]
  \centering
  \includegraphics[width=\linewidth]{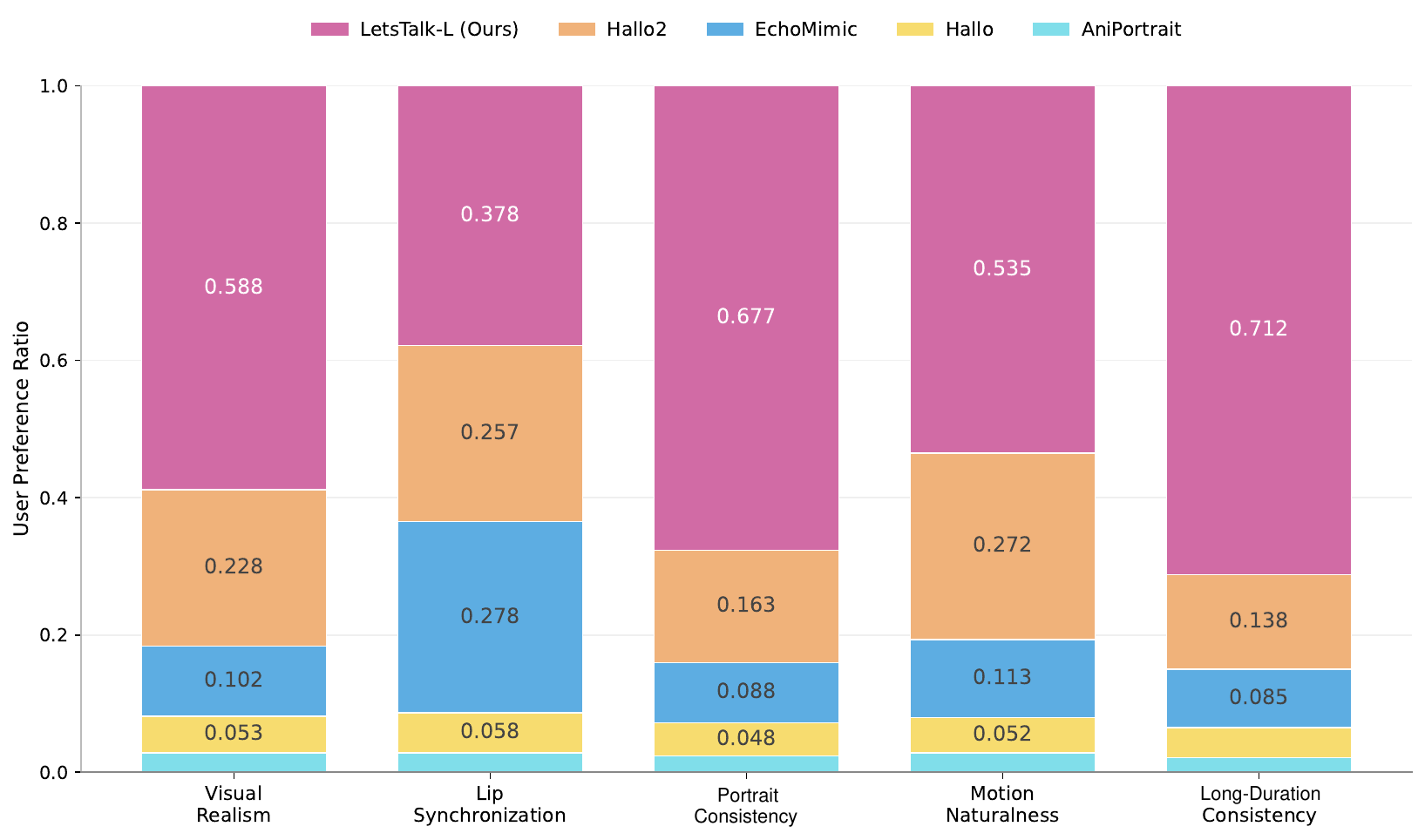}
  \caption{User preference evaluation results. We visualize the proportion of rankings where each method was preferred. LetsTalk-L (Ours) dominates across all metrics, particularly in portrait consistency and long-duration consistency.}
  \label{fig:user_study}
\end{figure}

\subsection{User Study}
We conducted a user study with 20 participants, comparing our method against Hallo~\cite{xu2024hallo}, EchoMimic~\cite{chen2024echomimic}, AniPortrait~\cite{wei2024aniportrait}, and Hallo2~\cite{cui2024hallo2}. Using 30 diverse audio samples, including long sequences $>10$s, participants ranked the models on five perceptual criteria: visual realism, lip synchronization, portrait consistency, motion naturalness, and long-duration consistency. The samples were shown in random order with hidden method names. Before evaluation, we provided brief written instructions for all five criteria and several non-test examples; participants could replay each sample before ranking. The participant pool included graduate students and researchers familiar with visual generation or multimedia analysis, as well as several general users, so that the study reflects both technical and perceptual preferences.
The preference rates are shown in Fig.~\ref{fig:user_study}. LetsTalk-L leads across all categories with a substantial margin. Notably, in long-duration consistency and portrait consistency, our method dominates with 71.2\% and 67.7\% respectively, significantly outperforming the runner-up Hallo2. This validates that our memory bank effectively eliminates flickering and portrait drift. While EchoMimic shows competitive performance in Lip Synchronization (27.8\%) and Hallo2 follows in Motion Naturalness (27.2\%), users overwhelmingly favored LetsTalk-L for its superior structural stability and visual realism. AniPortrait ranked lowest due to rigid poses and artifacts in extended generation.

% \begin{figure}[!t]
%   \centering
%   % \hspace{0.5cm}
%   \includegraphics[height=3.3cm]{Figure/Realism.pdf}
%   % \hspace{0.5cm}
%   \includegraphics[height=3.3cm]{Figure/Synchronization.pdf}
%   \caption{User study results (\%) on (a) realism (\emph{left}) and (b) synchronization (\emph{right}).}
%   \label{fig:user_study}
% \end{figure}

% \subsection{Visualization}
% We provide video visualizations showcasing long-duration results of up to 2 minutes. In these videos, each row corresponds to a different portrait driven by the same audio segment, demonstrating our model's ability to generate consistent and diverse talking head videos across various identities. Notably, our model is trained only on HDTF~\cite{zhang2021flow} and CelebV-HQ~\cite{zhu2022celebv}, which feature English conversational scenarios with real portraits. To further demonstrate the generalization ability of our approach, we also present results where real portraits are driven by Chinese speech and singing audio, even though such scenarios are not included in the training data.

\section{Conclusion}

In this work, we present \textit{LetsTalk}, a diffusion transformer framework for efficient and high-quality long-duration talking video synthesis under multimodal guidance. LetsTalk introduces a noise-regularized memory bank to maintain contextual continuity and mitigate error accumulation in extended video generation. Our spatiotemporal modeling, based on a deep compression autoencoder and linear attention, achieves state-of-the-art visual fidelity with substantially fewer parameters and faster inference. Through a systematic analysis of multimodal fusion, we show that Symbiotic Fusion for portrait features and Direct Fusion for audio provide strong portrait consistency and expressive, synchronized motion. Extensive experiments show that LetsTalk consistently outperforms existing approaches in quality, temporal coherence, and efficiency, while generalizing robustly to out-of-domain scenarios. Overall, LetsTalk advances realistic and scalable talking video generation, supporting broader applications in digital humans and interactive media.

\section*{Acknowledgment}

This work was supported by the National Natural Science Foundation of China (No. U2436210, No. 62306316). This work was completed while Haojie Zhang was a visiting student at Tsinghua University. Haojie Zhang, Zhihao Liang, Bingyan Liu, and Yaling Liang are with the South China University of Technology, Guangzhou, China; Ruibo Fu and Xuefei Liu are with the Institute of Automation, Chinese Academy of Sciences, Beijing, China (e-mail: \texttt{ruibofu@126.com}); Zhengqi Wen is with the Beijing National Research Center for Information Science and Technology, Tsinghua University, Beijing, China; Jianhua Tao is with the Department of Automation, BNRist, Tsinghua University, Beijing, China.

\bibliographystyle{IEEEtran}
\bibliography{IEEEabrv,./Reference}

\begin{thebibliography}{10}
\providecommand{\url}[1]{#1}
\csname url@samestyle\endcsname
\providecommand{\newblock}{\relax}
\providecommand{\bibinfo}[2]{#2}
\providecommand{\BIBentrySTDinterwordspacing}{\spaceskip=0pt\relax}
\providecommand{\BIBentryALTinterwordstretchfactor}{4}
\providecommand{\BIBentryALTinterwordspacing}{\spaceskip=\fontdimen2\font plus
\BIBentryALTinterwordstretchfactor\fontdimen3\font minus
  \fontdimen4\font\relax}
\providecommand{\BIBforeignlanguage}[2]{{%
\expandafter\ifx\csname l@#1\endcsname\relax
\typeout{** WARNING: IEEEtran.bst: No hyphenation pattern has been}%
\typeout{** loaded for the language `#1'. Using the pattern for}%
\typeout{** the default language instead.}%
\else
\language=\csname l@#1\endcsname
\fi
#2}}
\providecommand{\BIBdecl}{\relax}
\BIBdecl

\bibitem{zhang2023sadtalker}
W.~Zhang, X.~Cun, X.~Wang, Y.~Zhang, X.~Shen, Y.~Guo, Y.~Shan, and F.~Wang,
  ``{SadTalker: Learning Realistic 3D Motion Coefficients for Stylized
  Audio-Driven Single Image Talking Face Animation},'' in \emph{Proceedings of
  the IEEE/CVF Conference on Computer Vision and Pattern Recognition}, 2023,
  pp. 8652--8661.

\bibitem{wei2024aniportrait}
H.~Wei, Z.~Yang, and Z.~Wang, ``{AniPortrait: Audio-Driven Synthesis of
  Photorealistic Portrait Animation},'' \emph{arXiv preprint arXiv:2403.17694},
  2024.

\bibitem{wang2022anyonenet}
X.~Wang, Q.~Xie, J.~Zhu, L.~Xie, and O.~Scharenborg, ``Anyonenet: Synchronized
  speech and talking head generation for arbitrary persons,'' \emph{IEEE
  Transactions on Multimedia}, vol.~25, pp. 6717--6728, 2022.

\bibitem{ma2025talkclip}
Y.~Ma, S.~Wang, Y.~Ding, B.~Ma, T.~Lv, C.~Fan, Z.~Hu, Z.~Deng, and X.~Yu,
  ``Talkclip: Talking head generation with text-guided expressive speaking
  styles,'' \emph{IEEE Transactions on Multimedia}, 2025.

\bibitem{10.1145/311535.311556}
\BIBentryALTinterwordspacing
V.~Blanz and T.~Vetter, ``{A Morphable Model For The Synthesis Of 3D Faces},''
  in \emph{Proceedings of the 26th Annual Conference on Computer Graphics and
  Interactive Techniques}, ser. SIGGRAPH '99.\hskip 1em plus 0.5em minus
  0.4em\relax USA: ACM Press/Addison-Wesley Publishing Co., 1999, p. 187–194.
  [Online]. Available: \url{https://doi.org/10.1145/311535.311556}
\BIBentrySTDinterwordspacing

\bibitem{li2017learning}
T.~Li, T.~Bolkart, M.~J. Black, H.~Li, and J.~Romero, ``{Learning a model of
  facial shape and expression from 4D scans},'' \emph{ACM Trans. Graph.},
  vol.~36, no.~6, pp. 194--1, 2017.

\bibitem{bao2021high}
L.~Bao, X.~Lin, Y.~Chen, H.~Zhang, S.~Wang, X.~Zhe, D.~Kang, H.~Huang,
  X.~Jiang, J.~Wang \emph{et~al.}, ``{High-Fidelity 3D Digital Human Head
  Creation from RGB-D Selfies},'' \emph{ACM Transactions on Graphics (TOG)},
  vol.~41, no.~1, pp. 1--21, 2021.

\bibitem{chen2019hierarchical}
L.~Chen, R.~K. Maddox, Z.~Duan, and C.~Xu, ``{Hierarchical Cross-Modal Talking
  Face Generation With Dynamic Pixel-Wise Loss},'' in \emph{Proceedings of the
  IEEE/CVF conference on computer vision and pattern recognition}, 2019, pp.
  7832--7841.

\bibitem{prajwal2020lip}
K.~Prajwal, R.~Mukhopadhyay, V.~P. Namboodiri, and C.~Jawahar, ``{A Lip Sync
  Expert Is All You Need for Speech to Lip Generation in the Wild},'' in
  \emph{Proceedings of the 28th ACM international conference on multimedia},
  2020, pp. 484--492.

\bibitem{zhou2020makelttalk}
Y.~Zhou, X.~Han, E.~Shechtman, J.~Echevarria, E.~Kalogerakis, and D.~Li,
  ``{MakeItTalk: Speaker-Aware Talking-Head Animation},'' \emph{ACM
  Transactions On Graphics (TOG)}, vol.~39, no.~6, pp. 1--15, 2020.

\bibitem{cheng2022videoretalking}
K.~Cheng, X.~Cun, Y.~Zhang, M.~Xia, F.~Yin, M.~Zhu, X.~Wang, J.~Wang, and
  N.~Wang, ``{VideoReTalking: Audio-based lip synchronization for talking head
  video editing in the wild},'' in \emph{SIGGRAPH Asia 2022 Conference Papers},
  2022, pp. 1--9.

\bibitem{zhou2021pose}
H.~Zhou, Y.~Sun, W.~Wu, C.~C. Loy, X.~Wang, and Z.~Liu, ``{Pose-Controllable
  Talking Face Generation by Implicitly Modularized Audio-Visual
  Representation},'' in \emph{Proceedings of the IEEE/CVF conference on
  computer vision and pattern recognition}, 2021, pp. 4176--4186.

\bibitem{yi2022predicting}
R.~Yi, Z.~Ye, Z.~Sun, J.~Zhang, G.~Zhang, P.~Wan, H.~Bao, and Y.-J. Liu,
  ``Predicting personalized head movement from short video and speech signal,''
  \emph{IEEE Transactions on Multimedia}, vol.~25, pp. 6315--6328, 2022.

\bibitem{zhao2024ta2v}
M.~Zhao, W.~Wang, T.~Chen, R.~Zhang, and R.~Li, ``Ta2v: Text-audio guided video
  generation,'' \emph{IEEE Transactions on Multimedia}, vol.~26, pp.
  7250--7264, 2024.

\bibitem{ho2020denoising}
J.~Ho, A.~Jain, and P.~Abbeel, ``{Denoising Diffusion Probabilistic Models},''
  \emph{Advances in neural information processing systems}, vol.~33, pp.
  6840--6851, 2020.

\bibitem{dhariwal2021diffusion}
P.~Dhariwal and A.~Nichol, ``{Diffusion Models Beat Gans on Image Synthesis},''
  \emph{Advances in neural information processing systems}, vol.~34, pp.
  8780--8794, 2021.

\bibitem{rombach2022high}
R.~Rombach, A.~Blattmann, D.~Lorenz, P.~Esser, and B.~Ommer, ``{High-Resolution
  Image Synthesis With Latent Diffusion Models},'' in \emph{Proceedings of the
  IEEE/CVF conference on computer vision and pattern recognition}, 2022, pp.
  10\,684--10\,695.

\bibitem{zhang2023adding}
L.~Zhang, A.~Rao, and M.~Agrawala, ``{Adding Conditional Control to
  Text-to-Image Diffusion Models},'' in \emph{Proceedings of the IEEE/CVF
  International Conference on Computer Vision}, 2023, pp. 3836--3847.

\bibitem{ma2024latte}
X.~Ma, Y.~Wang, G.~Jia, X.~Chen, Z.~Liu, Y.-F. Li, C.~Chen, and Y.~Qiao,
  ``{Latte: Latent Diffusion Transformer for Video Generation},'' \emph{arXiv
  preprint arXiv:2401.03048}, 2024.

\bibitem{lu2023vdt}
H.~Lu, G.~Yang, N.~Fei, Y.~Huo, Z.~Lu, P.~Luo, and M.~Ding, ``{VDT:
  General-Purpose Video Diffusion Transformers via Mask Modeling},''
  \emph{arXiv preprint arXiv:2305.13311}, 2023.

\bibitem{hu2024animate}
L.~Hu, ``{Animate Anyone: Consistent and Controllable Image-To-Video Synthesis
  for Character Animation},'' in \emph{Proceedings of the IEEE/CVF Conference
  on Computer Vision and Pattern Recognition}, 2024, pp. 8153--8163.

\bibitem{xu2024hallo}
M.~Xu, H.~Li, Q.~Su, H.~Shang, L.~Zhang, C.~Liu, J.~Wang, L.~Van~Gool, Y.~Yao,
  and S.~Zhu, ``{Hallo: Hierarchical Audio-Driven Visual Synthesis for Portrait
  Image Animation},'' \emph{arXiv preprint arXiv:2406.08801}, 2024.

\bibitem{chen2024echomimic}
Z.~Chen, J.~Cao, Z.~Chen, Y.~Li, and C.~Ma, ``{EchoMimic: Lifelike Audio-Driven
  Portrait Animations through Editable Landmark Conditions},'' \emph{arXiv
  preprint arXiv:2407.08136}, 2024.

\bibitem{cui2024hallo2}
J.~Cui, H.~Li, Y.~Yao, H.~Zhu, H.~Shang, K.~Cheng, H.~Zhou, S.~Zhu, and
  J.~Wang, ``{Hallo2: Long-Duration and High-Resolution Audio-Driven Portrait
  Image Animation},'' \emph{arXiv preprint arXiv:2410.07718}, 2024.

\bibitem{xie2024sana}
E.~Xie, J.~Chen, J.~Chen, H.~Cai, H.~Tang, Y.~Lin, Z.~Zhang, M.~Li, L.~Zhu,
  Y.~Lu \emph{et~al.}, ``Sana: Efficient high-resolution image synthesis with
  linear diffusion transformers,'' \emph{arXiv preprint arXiv:2410.10629},
  2024.

\bibitem{ho2022video}
J.~Ho, T.~Salimans, A.~Gritsenko, W.~Chan, M.~Norouzi, and D.~J. Fleet,
  ``{Video Diffusion Models},'' \emph{Advances in Neural Information Processing
  Systems}, vol.~35, pp. 8633--8646, 2022.

\bibitem{guo2023animatediff}
Y.~Guo, C.~Yang, A.~Rao, Z.~Liang, Y.~Wang, Y.~Qiao, M.~Agrawala, D.~Lin, and
  B.~Dai, ``{Animatediff: Animate Your Personalized Text-To-Image Diffusion
  Models Without Specific Tuning},'' \emph{arXiv preprint arXiv:2307.04725},
  2023.

\bibitem{wang2024videocomposer}
X.~Wang, H.~Yuan, S.~Zhang, D.~Chen, J.~Wang, Y.~Zhang, Y.~Shen, D.~Zhao, and
  J.~Zhou, ``{VideoComposer: Compositional Video Synthesis With Motion
  Controllability},'' \emph{Advances in Neural Information Processing Systems},
  vol.~36, 2024.

\bibitem{chen2023videocrafter1}
H.~Chen, M.~Xia, Y.~He, Y.~Zhang, X.~Cun, S.~Yang, J.~Xing, Y.~Liu, Q.~Chen,
  X.~Wang \emph{et~al.}, ``{VideoCrafter1: Open Diffusion Models for
  High-Quality Video Generation},'' \emph{arXiv preprint arXiv:2310.19512},
  2023.

\bibitem{mou2024t2i}
C.~Mou, X.~Wang, L.~Xie, Y.~Wu, J.~Zhang, Z.~Qi, and Y.~Shan, ``T2i-adapter:
  Learning adapters to dig out more controllable ability for text-to-image
  diffusion models,'' in \emph{Proceedings of the AAAI Conference on Artificial
  Intelligence}, vol.~38, no.~5, 2024, pp. 4296--4304.

\bibitem{peebles2023scalable}
W.~Peebles and S.~Xie, ``{Scalable Diffusion Models With Transformers},'' in
  \emph{Proceedings of the IEEE/CVF International Conference on Computer
  Vision}, 2023, pp. 4195--4205.

\bibitem{ye2023geneface}
Z.~Ye, Z.~Jiang, Y.~Ren, J.~Liu, J.~He, and Z.~Zhao, ``{GeneFace: Generalized
  and High-Fidelity Audio-Driven 3D Talking Face Synthesis},'' \emph{arXiv
  preprint arXiv:2301.13430}, 2023.

\bibitem{stypulkowski2024diffused}
M.~Stypu{\l}kowski, K.~Vougioukas, S.~He, M.~Zieba, S.~Petridis, and M.~Pantic,
  ``{Diffused Heads: Diffusion Models Beat Gans on Talking-Face Generation},''
  in \emph{Proceedings of the IEEE/CVF Winter Conference on Applications of
  Computer Vision}, 2024, pp. 5091--5100.

\bibitem{ma2023dreamtalk}
Y.~Ma, S.~Zhang, J.~Wang, X.~Wang, Y.~Zhang, and Z.~Deng, ``{DreamTalk: When
  Emotional Talking Head Generation Meets Diffusion Probabilistic Models},''
  \emph{arXiv preprint arXiv:2312.09767}, 2023.

\bibitem{sun2023vividtalk}
X.~Sun, L.~Zhang, H.~Zhu, P.~Zhang, B.~Zhang, X.~Ji, K.~Zhou, D.~Gao, L.~Bo,
  and X.~Cao, ``{VividTalk: One-Shot Audio-Driven Talking Head Generation Based
  on 3D Hybrid Prior},'' \emph{arXiv preprint arXiv:2312.01841}, 2023.

\bibitem{tao2023galip}
M.~Tao, B.-K. Bao, H.~Tang, and C.~Xu, ``Galip: Generative adversarial clips
  for text-to-image synthesis,'' in \emph{Proceedings of the IEEE/CVF
  conference on computer vision and pattern recognition}, 2023, pp.
  14\,214--14\,223.

\bibitem{hou2025clip}
Y.~Hou, W.~Zhang, Z.~Zhu, and H.~Yu, ``Clip-gan: stacking clips and gan for
  efficient and controllable text-to-image synthesis,'' \emph{IEEE Transactions
  on Multimedia}, vol.~27, pp. 3702--3715, 2025.

\bibitem{tian2024emo}
L.~Tian, Q.~Wang, B.~Zhang, and L.~Bo, ``{EMO: Emote Portrait Alive-Generating
  Expressive Portrait Videos With audio2video Diffusion Model Under Weak
  Conditions},'' \emph{arXiv preprint arXiv:2402.17485}, 2024.

\bibitem{xu2024vasa}
S.~Xu, G.~Chen, Y.-X. Guo, J.~Yang, C.~Li, Z.~Zang, Y.~Zhang, X.~Tong, and
  B.~Guo, ``{VASA-1: Lifelike Audio-Driven Talking Faces Generated in Real
  Time},'' \emph{arXiv preprint arXiv:2404.10667}, 2024.

\bibitem{liu2025moee}
H.~Liu, W.~Sun, D.~Di, S.~Sun, J.~Yang, C.~Zou, and H.~Bao, ``Moee: Mixture of
  emotion experts for audio-driven portrait animation,'' in \emph{Proceedings
  of the IEEE/CVF Conference on Computer Vision and Pattern Recognition
  (CVPR)}, 2025, pp. 26\,222--26\,231.

\bibitem{cui2025hallo3}
J.~Cui, H.~Li, Y.~Zhan \emph{et~al.}, ``Hallo3: Highly dynamic and realistic
  portrait image animation with video diffusion transformer,'' in
  \emph{Proceedings of the IEEE/CVF Conference on Computer Vision and Pattern
  Recognition (CVPR)}, 2025, pp. 21\,086--21\,095.

\bibitem{ma2025tuning}
Y.~Ma, J.~Chen, D.~Di \emph{et~al.}, ``Tuning-free long video generation via
  global-local collaborative diffusion,'' \emph{arXiv preprint
  arXiv:2501.05484}, 2025.

\bibitem{qiu2024freenoise}
H.~Qiu, M.~Xia, Y.~Zhang, Y.~He, X.~Wang, Y.~Shan, and Z.~Liu, ``Freenoise:
  Tuning-free longer video diffusion via noise rescheduling,'' in
  \emph{Proceedings of the International Conference on Learning Representations
  (ICLR)}, 2024.

\bibitem{henschel2025streamingt2v}
R.~Henschel, L.~Khachatryan, H.~Poghosyan, D.~Hayrapetyan, V.~Tadevosyan,
  Z.~Wang, S.~Navasardyan, and H.~Shi, ``Streamingt2v: Consistent, dynamic, and
  extendable long video generation from text,'' in \emph{Proceedings of the
  Computer Vision and Pattern Recognition Conference}, 2025, pp. 2568--2577.

\bibitem{zhuang2024vlogger}
S.~Zhuang, K.~Li, X.~Chen, Y.~Wang, Z.~Liu, Y.~Qiao, and Y.~Wang, ``Vlogger:
  Make your dream a vlog,'' in \emph{Proceedings of the IEEE/CVF Conference on
  Computer Vision and Pattern Recognition (CVPR)}, 2024, pp. 8806--8817.

\bibitem{zhu2022label}
L.~Zhu and Y.~Yang, ``Label independent memory for semi-supervised few-shot
  video classification,'' \emph{IEEE Transactions on Pattern Analysis and
  Machine Intelligence}, vol.~44, no.~1, pp. 273--285, 2022.

\bibitem{vaswani2017attention}
A.~Vaswani, ``{Attention Is All You Need},'' \emph{Advances in Neural
  Information Processing Systems}, 2017.

\bibitem{katharopoulos2020transformers}
A.~Katharopoulos, A.~Vyas, N.~Pappas, and F.~Fleuret, ``Transformers are rnns:
  Fast autoregressive transformers with linear attention,'' in
  \emph{International conference on machine learning}.\hskip 1em plus 0.5em
  minus 0.4em\relax PMLR, 2020, pp. 5156--5165.

\bibitem{shen2023difftalk}
S.~Shen, W.~Zhao, Z.~Meng, W.~Li, Z.~Zhu, J.~Zhou, and J.~Lu, ``{DiffTalk:
  Crafting Diffusion Models for Generalized Audio-Driven Portraits
  Animation},'' in \emph{Proceedings of the IEEE/CVF Conference on Computer
  Vision and Pattern Recognition}, 2023, pp. 1982--1991.

\bibitem{schneider2019wav2vec}
S.~Schneider, A.~Baevski, R.~Collobert, and M.~Auli, ``{wav2vec: Unsupervised
  Pre-training for Speech Recognition},'' \emph{arXiv preprint
  arXiv:1904.05862}, 2019.

\bibitem{zhang2021flow}
Z.~Zhang, L.~Li, Y.~Ding, and C.~Fan, ``{Flow-guided One-shot Talking Face
  Generation with a High-resolution Audio-visual Dataset},'' in
  \emph{Proceedings of the IEEE/CVF Conference on Computer Vision and Pattern
  Recognition}, 2021, pp. 3661--3670.

\bibitem{zhu2022celebv}
H.~Zhu, W.~Wu, W.~Zhu, L.~Jiang, S.~Tang, L.~Zhang, Z.~Liu, and C.~C. Loy,
  ``{CelebV-HQ: A Large-scale Video Facial Attributes Dataset},'' in
  \emph{European conference on computer vision}.\hskip 1em plus 0.5em minus
  0.4em\relax Springer, 2022, pp. 650--667.

\bibitem{karras2019style}
T.~Karras, S.~Laine, and T.~Aila, ``{A Style-Based Generator Architecture for
  Generative Adversarial Networks},'' in \emph{Proceedings of the IEEE/CVF
  conference on computer vision and pattern recognition}, 2019, pp. 4401--4410.

\bibitem{chung2016out}
J.~S. Chung and A.~Zisserman, ``Out of time: automated lip sync in the wild,''
  in \emph{Asian conference on computer vision}.\hskip 1em plus 0.5em minus
  0.4em\relax Springer, 2016, pp. 251--263.

\bibitem{deng2019accurate}
Y.~Deng, J.~Yang, S.~Xu, D.~Chen, Y.~Jia, and X.~Tong, ``{Accurate 3D Face
  Reconstruction with Weakly-Supervised Learning: From Single Image to Image
  Set},'' in \emph{Proceedings of the IEEE/CVF conference on computer vision
  and pattern recognition workshops}, 2019, pp. 0--0.

\bibitem{rossler2018faceforensics}
A.~R{\"o}ssler, D.~Cozzolino, L.~Verdoliva, C.~Riess, J.~Thies, and
  M.~Nie{\ss}ner, ``{FaceForensics: A Large-scale Video Dataset for Forgery
  Detection in Human Faces},'' \emph{arXiv preprint arXiv:1803.09179}, 2018.

\bibitem{kingma2013auto}
D.~P. Kingma, ``{Auto-Encoding Variational Bayes},'' \emph{arXiv preprint
  arXiv:1312.6114}, 2013.

\end{thebibliography}

\clearpage

\appendix
\subsection{Preliminaries}
\label{subsec:preliminary}

\noindent\textbf{Diffusion Formulation} The diffusion process incrementally corrupts data $x_0$ with Gaussian noise: $q(x_t|x_0)=\mathcal{N}(x_t; \sqrt{\bar{\alpha}_t}x_0, (1-\bar{\alpha}_t)\mathbf{I})$,
where $\bar{\alpha}_t$ are predetermined hyperparameters. By reparameterizing, we obtain:
$x_{t} = \sqrt{\bar{\alpha}_t}x_0 + \sqrt{1-\bar{\alpha}_t}\epsilon_t$, where $\epsilon \sim \mathcal{N}(0, \mathbf{I})$, forming a Markov chain over $T$ steps.

Diffusion models are trained to learn the reverse process, modeled as: $p_\theta(x_{t-1}|x_t) = \mathcal{N}(\mu_\theta(x_t), \Sigma_\theta(x_t))$, with neural networks parameterizing $p_\theta$. This is optimized via the variational lower bound of $\log p(x_0)$:

\begin{equation}
\begin{split}
\mathcal{L}(\theta) = & \sum_t \mathcal{D}_{\text{KL}}(q^*(x_{t-1}|x_t,x_0) \,\|\, p_\theta(x_{t-1}|x_t))  \\ 
& - \log p_\theta(x_0|x_1).
\end{split}
\end{equation}

With both $q$ and $p_\theta$ Gaussian, $D_{\text{KL}}$ is computed from their means and covariances. Reparameterizing $\mu_\theta$ as a noise-prediction network $\epsilon_\theta$ enables training via mean-squared error between $\epsilon_\theta(x)$ and sampled noise $\epsilon_t$:

\begin{equation}
\mathcal{L}_{\text{simple}}(\theta) = \|\epsilon_\theta(x) - \epsilon(x)\|_2^2.
\end{equation}

To incorporate a learnable reverse covariance $\Sigma_\theta$, we minimize the full $D_{\text{KL}}$ term, following Nichol and Dhariwal: optimizing $\theta$ with $\mathcal{L}_{\text{simple}}$ and $\Sigma_\theta$ with the full $\mathcal{L}$.

\noindent\textbf{Diffusion Transformer}
To better handle multimodal fusion in the talking head task, we are motivated to extend the Diffusion Transformer (DiT) in our framework.
DiT builds upon the Vision Transformer (ViT) architecture, which processes sequences of image patches.
This sequential patch-based processing mechanism natively supports the fusion of multi-modality sufficiently and efficiently.
Different from ViT, DiT incorporates a variational autoencoder (VAE)~\cite{kingma2013auto} in the framework and therefore operates in latent space rather than pixel space.
This design enables DiT to handle high-resolution generation more efficiently, which is especially beneficial for video generation.

\subsection{Network Architecture Details}
\label{sec:network}

\subsubsection{Adaptive Integration} Following VDT\cite{lu2023vdt}, we insert an adaptive fully connected layer between attention layers:
\begin{equation}
    \bm{Z}_s = \bm{Z_s} + \alpha\cdot\text{FC}(\text{Temp-Attn}(\bm{Z_t}))
\end{equation}
where $\alpha$ is a learnable scalar and $\text{FC}(\cdot)$ denotes a fully-connected layer. This enables adaptive fusion of temporal dynamics into spatial representations.

\subsubsection{Audio Encoder}

We resample the audio signal to a 16 kHz sample rate. For audio feature extraction, we use wav2vec~\cite{schneider2019wav2vec}, employing embeddings from the last 12 layers to capture a richer range of semantic information across different levels of the audio representation. To ensure frame-by-frame alignment between audio and animation, we match the audio features with corresponding frames of the video clips. Additionally, we use a sliding window (multiple audio frames) for each video frame to maintain audio continuity and preserve more accurate audio information. To extract audio representations suitable for transformer tokenization, we designed an audio projection module. This module first flattens the extracted audio features and then selects a fixed number of context tokens (32 in our experiments). Simultaneously, the audio representations are mapped to the same dimensional feature space as the latent noise.

\subsubsection{Siamese Transformer}

The Siamese Fusion scheme, or ReferenceNet-like strategy, has been widely adopted in U-Net based approaches such as AnimateAnyone~\cite{hu2024animate}, EMO~\cite{tian2024emo}, and Hallo~\cite{xu2024hallo} to preserve identity features.
Typically, these methods employ a parallel U-Net encoder to extract hierarchical features and inject them into the main denoising U-Net block-wise.
Adapting this concept to our Diffusion Transformer framework, we design a specialized Parallel Transformer for the reference portrait image.
To ensure strictly aligned feature guidance, this parallel network mirrors the architectural configuration of our main denoising backbone (i.e., the Latte-based DiT~\cite{ma2024latte}). 
However, given that the reference portrait is a static image lacking temporal dynamics, we exclude the temporal attention modules from this parallel branch, retaining only the spatial transformer blocks. 
This design allows for efficient extraction of spatial appearance features, which are then fused into the main backbone to guide generation.
For the input audio, we propose a lightweight transformer where each block consists of a two-layer MLP to implement the Siamese Fusion scheme.

\begin{figure*}[!t]
\centering
\includegraphics[width=\linewidth]{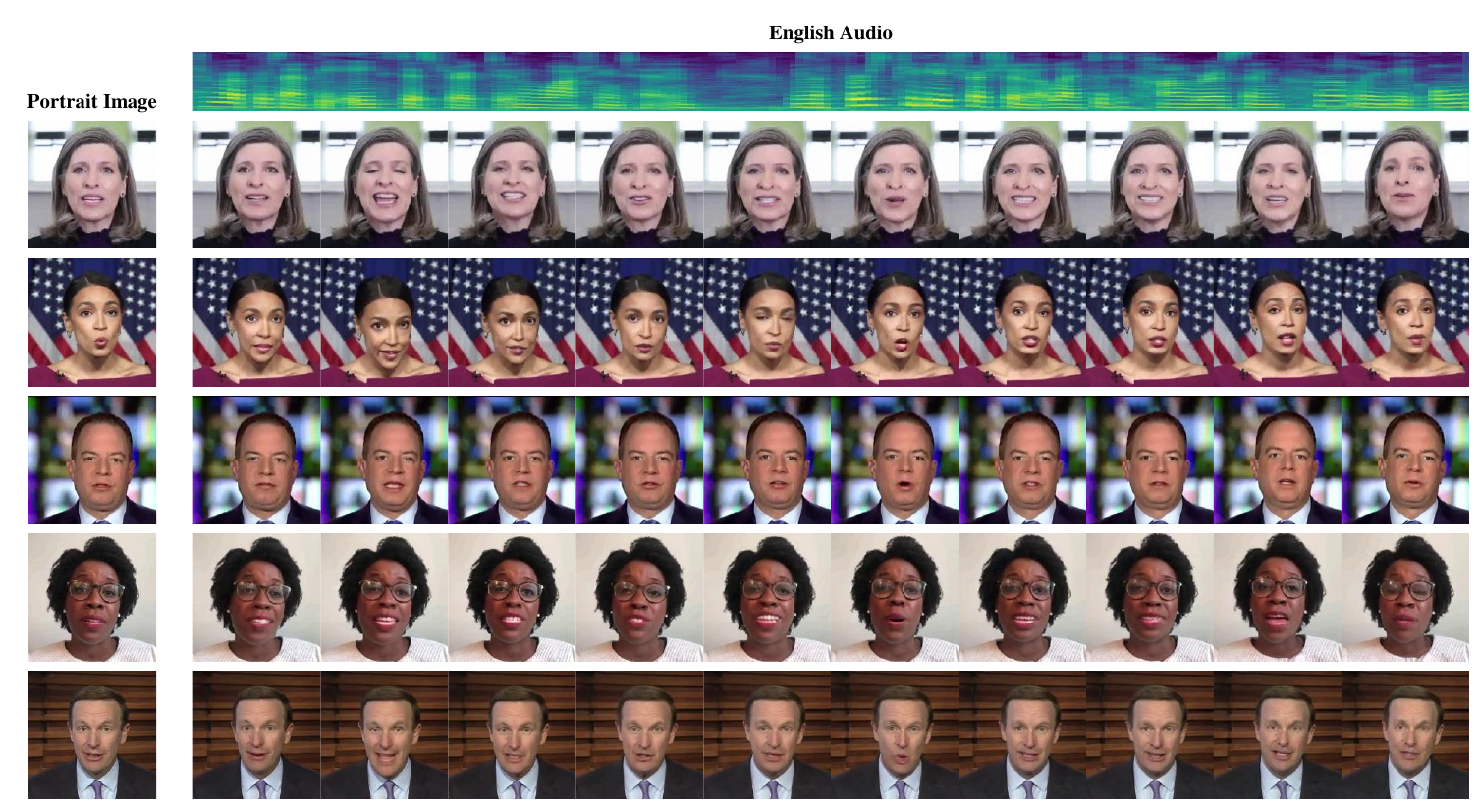}
\caption{English audio-driven results of LetsTalk. Note that each column corresponds to the same audio.}
\label{fig:English}
\end{figure*}

\begin{figure*}[!t]
\centering
\includegraphics[width=\linewidth]{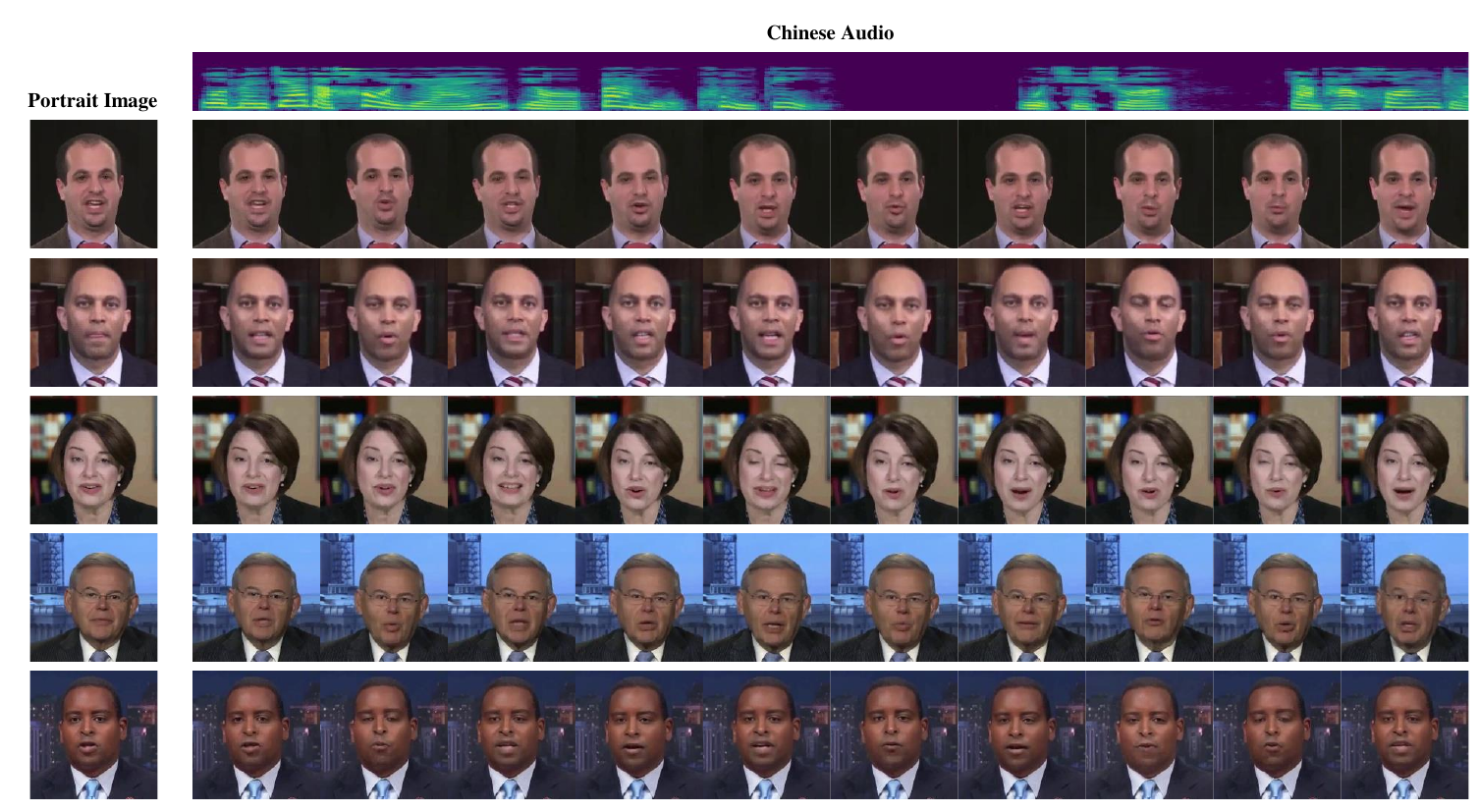}
\caption{Chinese audio-driven results of LetsTalk. Note that each column corresponds to the same audio.}
\label{fig:Chinese}
\end{figure*}

\begin{figure*}[!t]
\centering
\includegraphics[width=\linewidth]{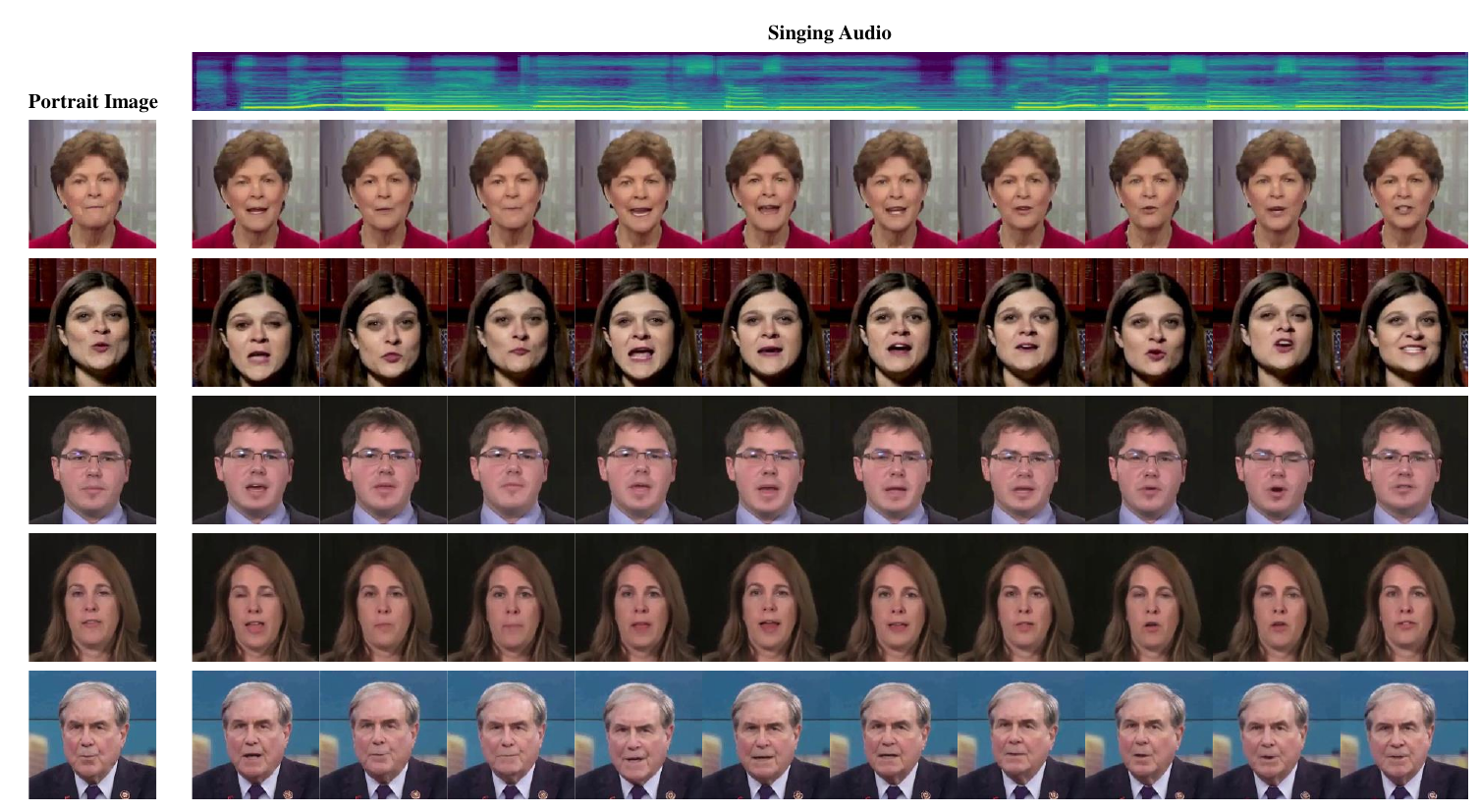}
\caption{Singing audio-driven results of LetsTalk. Note that each column corresponds to the same audio.}
\label{fig:Sing}
\end{figure*}

\subsection{Additional Qualitative Results}
\label{sec:qualitative}

This supplementary document further provides qualitative visualizations for long-duration generation, cross-domain generalization, and robustness under perturbed reference conditions. Together, these results complement the main paper by showing both the strength of LetsTalk in standard scenarios and its behavior under more challenging references.

\subsubsection{Long-Duration and Cross-Domain Visualizations}
\label{sec:visualization}

We provide video visualizations showcasing long-duration results of up to 2 minutes. In these videos, each row corresponds to a different portrait driven by the same audio segment, demonstrating our model's ability to generate consistent and diverse talking head videos across various identities. Notably, our model is trained only on HDTF~\cite{zhang2021flow} and CelebV-HQ~\cite{zhu2022celebv}, which feature English conversational scenarios with real portraits. To further demonstrate the generalization ability of our approach, we also present results where real portraits are driven by Chinese speech and singing audio, even though such scenarios are not included in the training data.

\subsubsection{Robustness under Perturbed Reference Conditions}
\label{sec:robustness}

We additionally provide four groups of qualitative comparisons under perturbed reference conditions, including head pose variation, harsh lighting, extreme expression, and partial occlusion. For each group, we fix the same driving audio and compare the generated results obtained from the original reference frame and the perturbed reference frame. In every visualization, the first column shows the reference frame, while the remaining columns present representative generated frames under the same audio. Rows correspond to different reference conditions, and frames in the same column are aligned to similar mouth states whenever possible, allowing direct comparison of portrait consistency, lip synchronization, and appearance changes caused by the perturbation.

\subsubsection{Impact of Reference Head Pose}

LetsTalk is most reliable when the reference portrait is close to the pose distribution seen during training, namely frontal or near-frontal views. When the reference image exhibits large yaw or pitch angles, the visible facial geometry becomes incomplete and the portrait tokens provide less reliable identity supervision. In such cases, the generated video may preserve coarse identity cues while losing fine-grained local details around the contour, cheeks, or far-side eye region. This limitation mainly stems from the reduced overlap between the static portrait prior and the target frames to be generated.

\begin{figure*}[!t]
\centering
\includegraphics[width=0.4\linewidth]{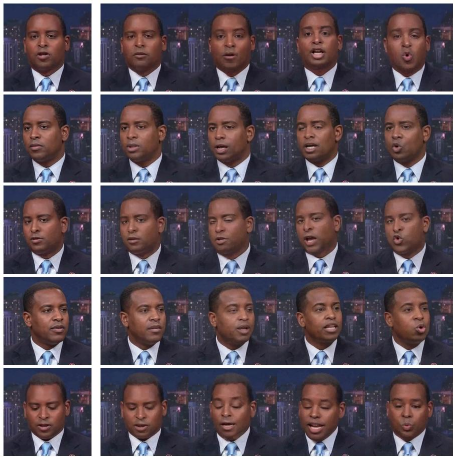}
\hspace{0.5em}
\includegraphics[width=0.4\linewidth]{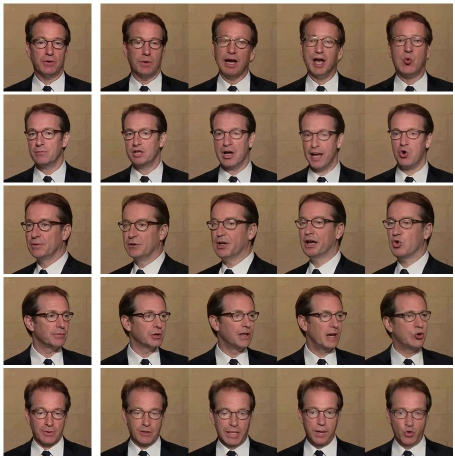}
\caption{Impact of reference head pose on generated results. We animate the same identity under different reference head poses while keeping the driving audio fixed. The comparison shows how pose variation in the reference frame affects portrait consistency, mouth synchronization, and local facial appearance in the final talking-head results.}
\label{fig:supp_head_pose}
\end{figure*}

\subsubsection{Robustness to Harsh Lighting}

The method is reasonably robust to moderate illumination variation, but harsh lighting remains challenging. Strong backlighting, high-contrast shadows, or saturated highlights in the reference frame can be encoded into the portrait features and propagated through the denoising process, which may lead to unstable skin texture recovery or local brightness inconsistency across frames. The visualization helps illustrate that the model remains broadly stable, while severe lighting perturbations may still induce local appearance shifts.

\begin{figure*}[!t]
\centering
\includegraphics[width=0.4\linewidth]{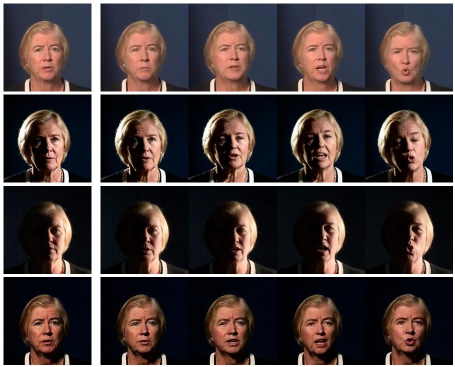}
\hspace{0.5em}
\includegraphics[width=0.4\linewidth]{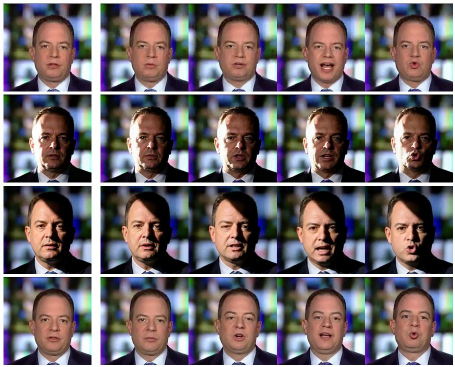}
\caption{Robustness to harsh lighting in the reference frame. We introduce extreme illumination conditions, such as strong side lighting or top lighting, into the reference portrait and compare the generated videos with those obtained from the original reference.}
\label{fig:supp_harsh_lighting}
\end{figure*}

\subsubsection{Influence of the Reference Expression}
Because the reference portrait provides strong static appearance guidance, its expression can affect the generated result, especially when the reference contains exaggerated smiles, tightly closed eyes, or other extreme facial states. In those cases, the model may partially inherit expression-specific texture cues that interfere with fully neutral portrait consistency, making some generated frames appear slightly biased toward the reference expression.

\subsubsection{Impact of Partial Occlusions}
Partial occlusions in the reference portrait, such as microphones, hair crossing the face, or other foreground objects, remain a source of difficulty. Since the portrait branch and memory-guided generation rely on visible appearance cues, structured occlusions can introduce local artifact accumulation or identity ambiguity in the corresponding regions. The visualization highlights the overall stability of the model as well as the remaining sensitivity to severe occlusions.

\subsection{Discussion and Future Work}
\label{sec:limitations}

LetsTalk achieves strong overall performance, but several challenges remain under difficult reference conditions and over long temporal horizons. This section discusses these limitations and outlines possible directions for future improvement.

\subsubsection{Challenging Reference Conditions}
The robustness comparisons suggest that different perturbations affect the model in distinct ways. Large head pose variations reduce the completeness of visible facial geometry and weaken identity guidance in side regions. Harsh lighting can introduce local brightness inconsistency or unstable skin texture, since illumination patterns may be encoded together with identity cues. Extreme reference expressions may bias the generated appearance toward expression-specific textures, indicating that identity and expression are not fully disentangled in the portrait branch. Structured partial occlusions, such as microphones or other foreground objects, remain difficult because they corrupt the appearance cues used for portrait conditioning and may cause local ambiguity in the generated frames.

\begin{figure*}[!t]
\centering
\includegraphics[width=0.4\linewidth]{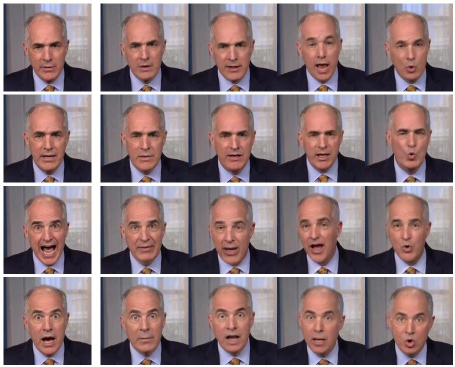}
\hspace{0.5em}
\includegraphics[width=0.4\linewidth]{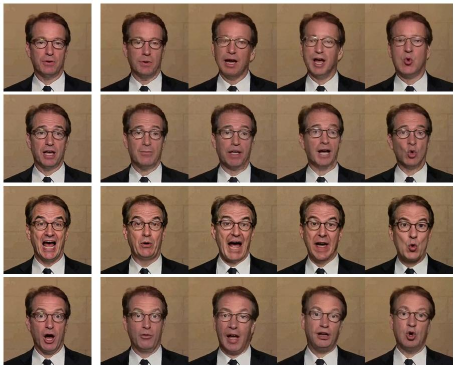}
\caption{Influence of reference expression on generated results. We replace the original reference portrait with versions exhibiting stronger speaking or exaggerated expressions, and analyze how expression intensity in the reference is transmitted to the final animation under the same audio condition.}
\label{fig:supp_extreme_expression}
\end{figure*}

\begin{figure*}[!t]
\centering
\includegraphics[width=0.4\linewidth]{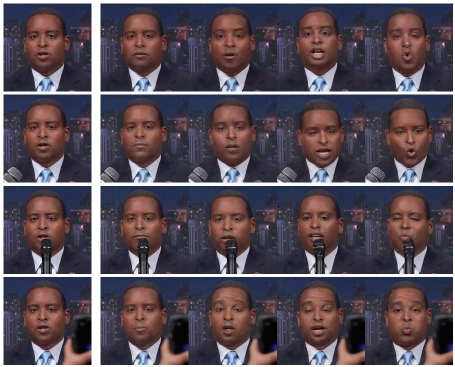}
\hspace{0.5em}
\includegraphics[width=0.4\linewidth]{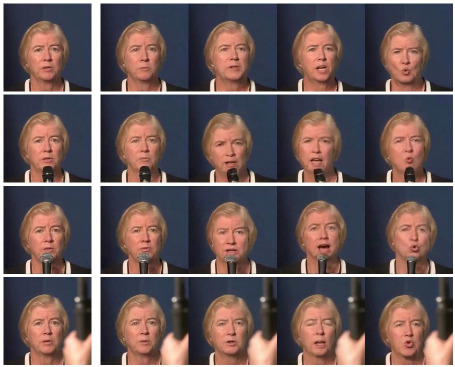}
\caption{Impact of partial occlusions in the reference frame. We add microphone-like foreground objects or other partial occlusions to the reference portrait and compare the resulting generations with those from the original reference.}
\label{fig:supp_partial_occlusion}
\end{figure*}

\subsubsection{Long-Duration Consistency}
The proposed noise-regularized memory bank improves long-duration consistency, but stability remains imperfect in challenging scenarios. Fast head motion, abrupt phoneme transitions, lighting variation, or occlusion changes can still introduce subtle flickering, texture softness, or short-term drift. These effects become more noticeable over long horizons, where small local errors may accumulate despite memory retrieval and regularization.

\subsubsection{Future Work}
Several directions may further improve the robustness and controllability of LetsTalk. A first direction is to design more robust portrait representations that are less sensitive to pose variation, illumination changes, extreme expressions, and partial occlusions. A second direction is to develop more selective memory retrieval and update strategies, so that unreliable historical cues are less likely to accumulate over long sequences. A third direction is to encourage stronger disentanglement of identity, expression, pose, and prosody, which may improve controllability while preserving stable identity and natural motion.

\end{document}